\documentclass[runningheads]{llncs}

\usepackage{eccv}

\usepackage{eccvabbrv}

\usepackage{graphicx}
\usepackage{booktabs}
\usepackage{wrapfig}
\usepackage{multirow}

\usepackage[accsupp]{axessibility}  

\usepackage{hyperref}

\usepackage{orcidlink}

\usepackage{pifont}
\newcommand{\cmark}{\ding{52}}  
\newcommand{\xmark}{\ding{56}} 

\begin{document}

\title{EgoSim: Egocentric World Simulator for Embodied  Interaction Generation} 

\titlerunning{EgoSim}

\author{Jinkun Hao\inst{1}$^{*}$ \and
Mingda Jia\inst{2}$^{*}$ \and
Ruiyan Wang\inst{1} \and
Hongrui Zhu\inst{2} \and
Jiafei Cao\inst{2} \and \\
Xihui Liu\inst{3} \and 
Ran Yi\inst{1}$^{\dagger}$ \and 
Lizhuang Ma\inst{1}$^{\dagger}$ \and
Jiangmiao Pang\inst{2} \and
Xudong Xu\inst{2}}

\authorrunning{J.~Hao et al.}

\institute{Shanghai Jiaotong University 
 \and
Shanghai AI Laboratory 
 \and
The University of Hong Kong\\
[1ex]
{\small $^{*}$Equal contribution \quad $^{\dagger}$Corresponding author}}

\maketitle

\begin{abstract}

We introduce EgoSim, a closed-loop egocentric world simulator that generates spatially consistent interaction videos and persistently updates the underlying 3D scene state for continuous simulation. Existing egocentric simulators either lack explicit 3D grounding, causing structural drift under viewpoint changes, or treat the scene as static, failing to update world states across multi-stage interactions. 
EgoSim addresses both limitations by modeling 3D scenes as updatable world states. We generate embodiment interactions via a Geometry-action-aware Observation Simulation model, with spatial consistency from an Interaction-aware State Updating module.
To overcome the critical data bottleneck posed by the difficulty in acquiring densely aligned scene–interaction training pairs, we design a scalable pipeline that extracts static point clouds, camera trajectories, and embodiment actions from in-the-wild large-scale monocular egocentric videos.
Extensive experiments demonstrate that EgoSim significantly outperforms existing methods in terms of visual quality, spatial consistency, and generalization to complex scenes and in-the-wild dexterous interactions, while supporting cross-embodiment transfer to robotic manipulation. Project page is at: \url{egosimulator.github.io}.

\keywords{Egocentric World Simulator \and World Models \and Generative Models}
\end{abstract}
\section{Introduction}
\label{sec:intro}

\begin{figure*}[t]
    \centering
    \includegraphics[width=\textwidth]{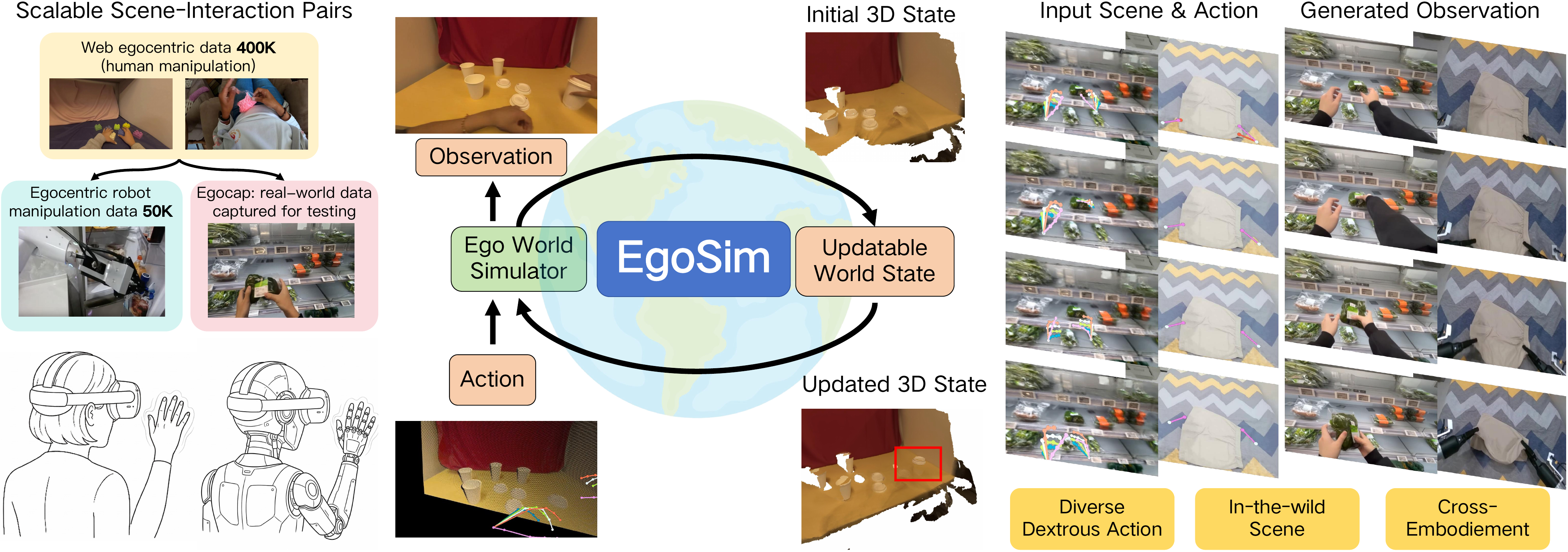}
    \caption{Given a scene image and a sequence of actions, \textbf{EgoSim} generates temporally and spatially consistent egocentric observations and high-quality dexterous interactions. Egosim also persistently updates the 3D scene state for continuous simulation. We propose a data construction pipeline to leverage web-scale egocentric video data, and thus strengthen the generalization ability of Egosim with scalable scene-interaction pairs. EgoSim also exhibits strong few-shot adaptation capability to real-world scenarios and diverse robotic embodiments.}
    \label{fig:teaser}
\end{figure*}

The advances in video diffusion models~\cite{wan2025,yang2024cogvideox,kong2024hunyuanvideo} have significantly advanced the development of interactive world simulators~\cite{genie_deepmind,robbyantteam2026advancingopensourceworldmodels,he2025matrix,feng2024matrix,mao2025yume},
revolutionizing how we model and interact with virtual environments, which demonstrates strong potential for applications in spatial intelligence, game engines, and embodied AI.
Although existing world simulators can generate realistic videos and construct immersive scenes, they only support restricted forms of action control like navigation, and typically render scenes from a third-person perspective, reducing users to passive spectators.
In contrast, humans and robots naturally function as active participants capable of performing far more diverse and flexible behaviors, such as fine-grained hand movements.

Recent studies~\cite{tu2025playeroneegocentricworldsimulator,wang2026hand2worldautoregressiveegocentricinteraction,kim2025dexterousworldmodels} have begun to explore egocentric world simulators that support vivid interactions driven by human motions. Given an egocentric scene image or video as the world context, these methods typically take full-body or hand-only actions as conditions and synthesize corresponding interactive videos for world simulation.
Despite achieving promising generation quality, such simulators still suffer from several critical limitations.

Some works~\cite{tu2025playeroneegocentricworldsimulator,wang2026hand2worldautoregressiveegocentricinteraction} leverage the implicit camera motion injection mechanism employed in video diffusion models, yet struggle to guarantee the 3D consistency of interaction videos and strict alignment between camera trajectories and generated frames.
To circumvent this hurdle, DWM~\cite{kim2025dexterousworldmodels} explicitly decouples static 3D scenes from the dynamic changes induced by user actions.
However, all of them ignore or fail to maintain a fundamental capability of any world simulator, {consistent world state updating}, especially during long-sequence generation.
Furthermore, current world simulators are only trained on restricted data with aligned multiview videos~\cite{tu2025playeroneegocentricworldsimulator,guo2023ft,grauman2024ego} or data collected from synthetic environments~\cite{kim2025dexterousworldmodels,jiang2024scaling}, which are orders of magnitude smaller than web-scale monocular egocentric videos. Undoubtedly, this severely limits the scalability of such models and consequently compromises their generalization ability.

In this paper, we present \textbf{EgoSim}, an egocentric world simulator as shown in Fig~\ref{fig:teaser} that models 3D static scenes as \emph{updatable} world states.
Inspired by DWM~\cite{kim2025dexterousworldmodels}, we also disentangle static 3D scenes from dynamic changes and formulate EgoSim as an ego-motion-action-aware observation simulation model generating egocentric interaction videos.
In contrast to prior work, we first take the initial egocentric frame, remove hands via inpainting, and further reconstruct the point cloud of static scenes to construct the spatial condition of Egosim.
Thanks to the editable nature of point clouds, EgoSim can conveniently memorize and update a persistent world state immediately after generating interaction observations from input actions, enabling continuous world simulation with supervision from a spatial-temporally persistent scene state. Specifically, we identify objects interacting with human or robotic arms and update their corresponding point clouds based on their latest observation changes. 
In contrast to existing approaches that either reconstruct only static scenes~\cite{zhao2025spatiavideogenerationupdatable} or simply track moving objects~\cite{huang2025vipe}, we first propose an Interaction-aware State Updating module that enables handling a wider range of interaction scenarios, including the manipulation of fixed articulated objects and multi-part assembly tasks. 

To enable the scaling of training data, we propose an elegant data construction pipeline for processing massive in‑the‑wild monocular egocentric videos.
Notably, EgoSim inherently requires \emph{well‑aligned paired data}, including a static 3D scene, camera trajectory, hand motion representations, and the egocentric video itself as the training target.
Accordingly, given a monocular egocentric video, our pipeline first obtains the static 3D scene using the aforementioned procedure.
We then leverage DepthAnything3~\cite{depthanything3} to estimate camera trajectories from the video, and employ HaMeR~\cite{pavlakos2024reconstructing} to extract hand keypoint sequences as motion conditions of EgoSim.
It is noteworthy that we adopt hand keypoints rather than hand meshes, as this choice facilitates the adaptation of our model to any embodiment, such as robotic grippers.

Through extensive experiments, we validate the effectiveness of our proposed explicit and updatable world states.
Given a scene image and dexterous hand actions, EgoSim not only synthesizes realistic interactive videos with great spatial consistency but also successfully updates 3D scenes to maintain consistent world model transitions.
Thanks to our well-designed data construction pipeline, EgoSim can be trained on web-scale monocular egocentric videos and thus exhibits strong generalization ability.
Specifically, EgoSim performs favorably in complex scenes with multiple interactive objects and is capable of generating diverse, sophisticated manipulations, including deformable object handling and small object assembling, which are unattainable by existing approaches.
Moreover, EgoSim demonstrates strong adaptation capability. After finetuning only 100 steps on AgiBot-World data~\cite{bu2025agibot}, it achieves promising performance on robotic arm manipulation~\cite{wang2026evaaligningvideoworld}.
We further design EgoCap, a low-cost data collection device, and collect 50 hand interaction video clips in real-world application scenarios.
Experimental results demonstrate that EgoSim can rapidly adapt to these real-world interactive videos.

\section{Related Work}
\label{sec:related}

\subsection{Interactive Video Generation}
The rapid progress of video diffusion models~\cite{wan2025,yang2024cogvideox,kong2024hunyuanvideo} has shifted world modeling from latent-space prediction~\cite{ha2018world} to high-fidelity visual simulation. Genie~\cite{genie_deepmind}, The Matrix~\cite{he2025matrix,feng2024matrix}, and Lingbot-World~\cite{robbyantteam2026advancingopensourceworldmodels} generate controllable video streams for game environments and embodied AI. However, these methods support only coarse controls like directional commands or camera poses. 

Meanwhile, several methods explore hand-object interaction 
video generation with finer-grained control signals. InterDyn~\cite{akkerman2025interdyn} conditions on binary hand masks via a ControlNet-like branch to probe the implicit physical knowledge of pretrained video models. CosHand~\cite{sudhakar2024coshand} generates single-frame hand-object interactions from hand mask inputs. Mask2IV~\cite{li2026mask2iv} introduces a two-stage pipeline that first predicts interaction mask trajectories and then synthesizes trajectory-conditioned videos. SpriteHand~\cite{li2025spritehand} renders hands over static backgrounds with autoregressive generation for real-time interaction. However, these methods operate on monocular 2D signals without explicit 3D scene grounding, and none of them maintains a persistent world state for continuous simulation.

\subsection{Egocentric World Simulators}
\label{sec:ego_sim}
Egocentric world simulators generate action-conditioned videos from a first-person perspective. PlayerOne~\cite{tu2025playeroneegocentricworldsimulator} conditions on whole-body motion derived from synchronized ego-exocentric captures. Hand2World~\cite{wang2026hand2worldautoregressiveegocentricinteraction} distills a bidirectional video diffusion model into a causal autoregressive generator for monocular streaming synthesis. Both methods encode the scene state implicitly without explicit 3D representations, which limits spatial consistency under large viewpoint changes. Moreover, PlayerOne relies on synchronized ego-exocentric video pairs for training, which are difficult to scale. DWM~\cite{kim2025dexterousworldmodels} decouples the static 3D scene from action-induced dynamics by conditioning on rendered point maps and hand meshes, thereby improving spatial consistency. However, its scene is reconstructed once and never updated after interactions, and its training relies on synthetic environments or paired captures that are equally difficult to scale. EgoSim maintains an explicit, updatable scene state, enabling precise condition-following ability for both action and geometric inputs, even under large-scale and drastic view changes.

\subsection{World Models with Scene States}
A fundamental requirement for a world simulator is to memorize and update the environment state from its generated
observations. Recent 3D reconstruction methods~\cite{depthanything3,wang2025moge2accuratemonoculargeometry, wu2024recent, li2026joint, lin2025review} can recover dense geometry from single-view images. Several works have explored updating representations by considering object dynamics in video sequences. VIPE~\cite{huang2025vipe} fuses per-frame point clouds through decoupling moving objects to maintain a clean static scene. Spatia~\cite{zhao2025spatiavideogenerationupdatable} discovers utilizing motion-aware scene states as geometric prior to enhance video generation. However, they overlook other object-embodiment interactions that are more complex than simple motion. WristWorld~\cite {qian2025wristworldgeneratingwristviews4d} reconstructs coarse 4D scene points for enhancing the spatial consistency of robot world models. In contrast, EgoSim performs fine-grained interaction-aware state sensing, which explicitly tracks and updates complex object interactions. This interaction-aware state serves as a more appropriate spatial prior for interactive world simulators.

\section{Method}
\label{sec:method}

\begin{figure}[t]
    \centering
    \includegraphics[width=\linewidth]{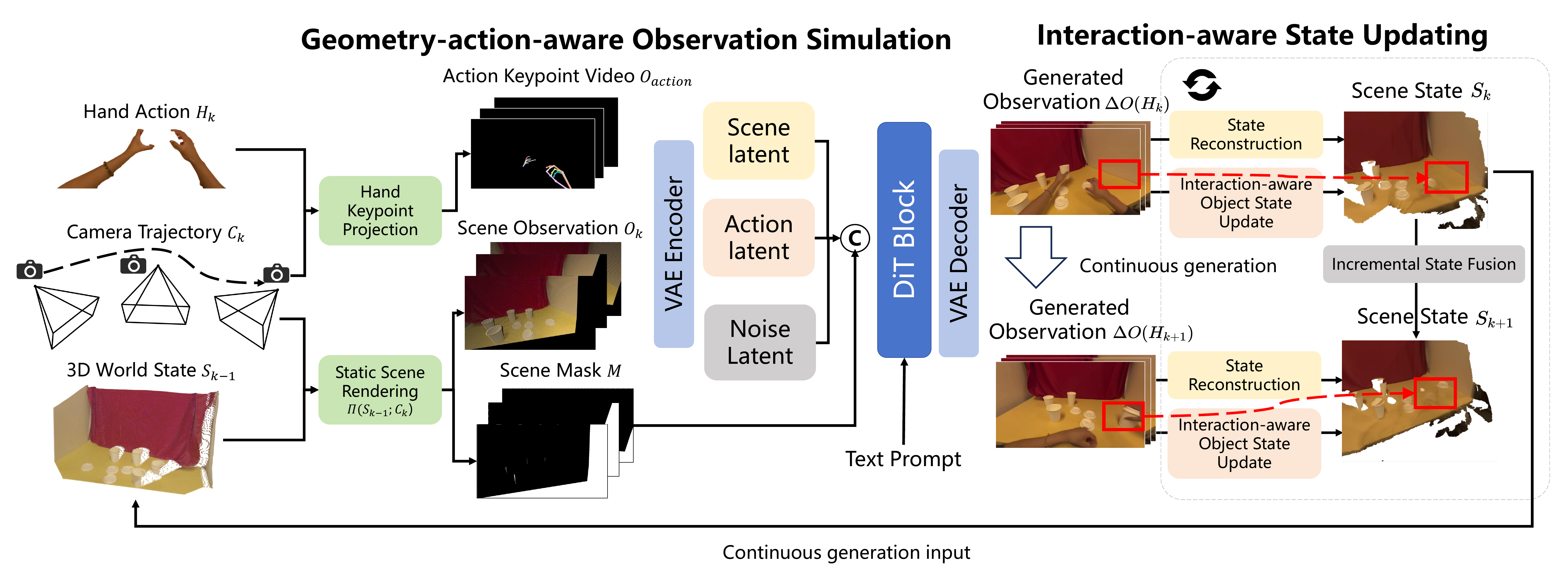}
    \caption{Overview of {EgoSim}. Our framework enables continuous egocentric simulation via updatable world modeling. It consists of (1) a {Geometry-action-aware Observation Simulation} model that synthesizes action-conditioned visual dynamics $O_k$ based on current scene geometry, and (2) an Interaction-aware State Updating module that tracks point-based interaction-aware object states from generated observations and integrates them back into the persistent 3D world state to update $S_k$.
    }
    \label{fig:Framwork}
\end{figure}

EgoSim operates as a closed-loop world simulator that cyclically alternates between visual observation generation and 3D state updates to ensure temporal and physical consistency, as illustrated in Figure~\ref{fig:Framwork}. In this section, we first formalize the simulation process (Sec.~\ref{sec:formulation}). We then detail its two core components. The Geometry-action-aware Observation Simulation model is responsible for state transitions (Sec.~\ref{sec:video_model}) and the Interaction-aware State Updating for long-term state persistence (Sec.~\ref{sec:memory}).
 
\subsection{Problem Formulation}
\label{sec:formulation}
We formulate the continuous world simulator by three core components: the environment \textbf{State ($S$)}, the interaction \textbf{Action ($A$)}, and the visual \textbf{Observation ($O$)}. 

At any given stage $k$, the simulator maintains a 3D world state $S_{k-1}$ representing the current scene state. The input action $A_k$ is explicitly decoupled into a camera trajectory $C_k$ and a hand interaction sequence $H_k$, expressed as $A_k = (C_k, H_k)$.

The simulation loop consists of two sequential operations. First, the model generates the intermediate visual observation $O_k$ by rendering the static background $\Pi(S_{k-1}; C_k)$ and synthesizing the dynamic observation residuals $\Delta O(H_k)$ induced by the hand action:
\begin{equation}
O_k = \Pi(S_{k-1}; C_k) + \Delta O(H_k)
\end{equation}
However, pure decoupled generation cannot preserve states across stages. To close the simulation loop, the system executes a 3D state update function $\mathcal{U}$ that extracts the latest physical layout from the generated observation $O_k$ and persistently applies it back to the 3D scene:
\begin{equation}
S_k = \mathcal{U}(S_{k-1}, O_k)
\end{equation}
By performing this localized state update, we prevent manipulated objects (e.g., an opened door) from reverting to their original layouts in subsequent renderings. This formulation naturally motivates our two core modules include the Observation Simulation Model (Sec.~\ref{sec:video_model}) and the interaction-aware State Updating (Sec.~\ref{sec:memory}).

\subsection{Geometry-action-aware Observation Simulation}
\label{sec:video_model}
The visual state transition is implemented through a geometry-action-aware video generation model acting as our Observation Simulation model. This model synthesizes the dynamic residuals driven by hand interactions, while ensuring strict adherence to the 
3D environment.

Furthermore, to rigorously constrain the generation process, we formulate the conditioning signals by combining explicit scene geometry with action representations. The {Scene Observation video} $O_{k}$ provides the background and accurate camera motion cues. Simultaneously, the interaction action is represented by a {hand keypoint video} $O_{action}$. Specifically, we extract 3D hand keypoints and project them onto the 2D observation plane using the corresponding camera intrinsics and extrinsics. This perspective projection enforces depth-dependent foreshortening and accurate geometry, anchoring the generated interactions in 3D space. We intentionally employ core joint keypoints rather than dense hand meshes, as keypoints serve as a universal, cross-embodiment action representation that facilitates generalization from human hands to robotic end-effectors. Furthermore, because $O_{k}$ inevitably contains {unrendered regions} due to occlusions or incomplete scanning, we introduce a {binary mask video} $M$ to explicitly indicate these unobserved regions that require visual synthesis.

\textbf{Training \& Inpainting Prior.} During training, we operate in the latent space of a pretrained video VAE. The denoising network $\epsilon_\theta$, instantiated as a Diffusion Transformer (DiT), processes the concatenated latent $z_{in}^{(t)} = \text{Concat}(z_t, z_{bg}, z_{hand}, M)$ to predict the Gaussian noise $\epsilon$, where all conditions are spatially aligned with the noisy latent $z_t$:
\begin{equation}
\mathcal{L}_{\text{gen}} = \mathbb{E}_{z_0, t, \epsilon \sim \mathcal{N}(\mathbf{0}, \mathbf{I})} \left[ \left\| \epsilon - \epsilon_\theta \big(z_{in}^{(t)}, t \big) \right\|_2^2 \right]
\end{equation}
We further initialize the DiT with pre-trained inpainting weights to address incomplete observations by hallucinating unmodeled regions in $M$. Furthermore, it acts as an {identity function with a generative prior}~\cite{kim2025dexterousworldmodels}, encouraging the model to preserve the known background $O_{k}$ and only activate generation in regions conditioned by actions in $O_{action}$.
\begin{figure}[t]
    \centering
    \includegraphics[width=0.95\linewidth]{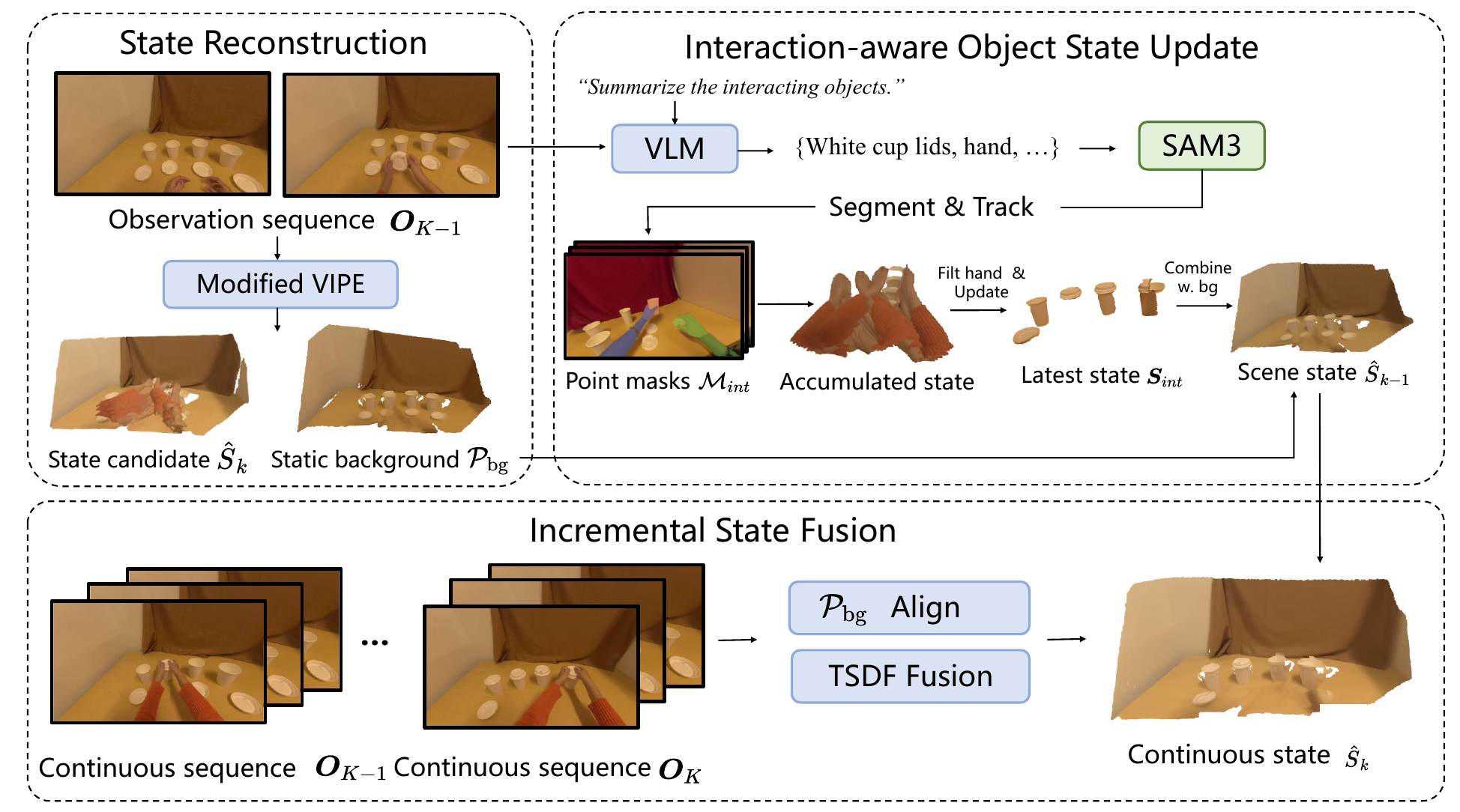}
    \caption{\textbf{Implementation details of Interaction-aware State Updating.} State Reconstruction (left) builds a point cloud from observations via modified VIPE~\cite{huang2025vipe}. Object State Update (top-right) identifies and tracks interactive objects with a VLM and SAM3, compositing their latest geometry into the scene. Incremental State Fusion (bottom) aligns and merges consecutive states via TSDF fusion.}
    \label{fig:memory_details}
\end{figure}

\subsection{Interaction-aware State Updating}
\label{sec:memory}

\subsubsection{State Reconstruction.}

To maintain a consistent 3D scene with updatable object states for continuous simulation, we design a training-free module to reconstruct the state candidate \( \hat{S}_k \) by estimating and fusing segment-level point clouds from the \( K \)-th incoming observation sequence \( \{O_0, O_1, \dots, O_{n-1}\} \subset \boldsymbol{O}_{K-1} \) with \( n \) frames, as an editable 3D environment representation. Inspired by VIPE~\cite{huang2025vipe}, our pipeline first estimates camera intrinsics via DepthAnything3~\cite{depthanything3} and extracts instance masks via SAM3~\cite{carion2025sam3segmentconcepts} guided by text phrases of interaction-related objects. Per-frame depths and camera poses are predicted and aligned using a dual-pass DROID-SLAM with multi-view depth alignment to resolve scale ambiguity. Building on this, the pipeline detects interactive objects via Grounding DINO~\cite{liu2023grounding} and SAM3~\cite{carion2025sam3segmentconcepts}, and decouples them from static objects during reconstruction.

\subsubsection{Interaction-aware Object State Update.}

To register and update the dynamic states \( \boldsymbol{S}_{int} \) of interactive objects, we first employ a vision-language model~\cite{qwen2.5} to identify object phrases involved in embodiment interactions. An open-vocabulary tracking pipeline based on SAM3~\cite{carion2025sam3segmentconcepts} then estimates a 3D point mask \( \mathcal{M}_{int}^i \) for each interactive object \( S_{int}^{i} \in \boldsymbol{S}_{int} \) through a hierarchical filtering process, we tag embodiment-centric instances as subject references, select mask candidates by IoU overlap, and refine them with depth consistency to suppress false positives. During reconstruction, point clouds excluding \( \mathcal{M}_{int} \) are fused into a static background \( \mathcal{P}_{\text{bg}} \), while interactive objects are tracked across frames and their latest-frame geometry is composited into \( \mathcal{P}_{\text{bg}} \) to form the final \( \hat{S}_k \).

\subsubsection{Incremental State Fusion.}

During continuous generation, the scene state \( S_k \) of generated observation sequence $\boldsymbol{O}_{K}$ is obtained by incrementally fusing \( S_{k-1} \) and \( \hat{S}_k \). We first align the coordinate systems of the two states via the Sim3 Umeyama algorithm computed from the overlapping camera poses \( C_k \) and \( C_{k-1} \), then apply TSDF fusion to merge their point clouds, with overlapping regions updated from \( \hat{S}_k \). Interactive objects retain their geometry from the last observed frame, yielding an up-to-date scene representation.
\section{Scalable Data Pipeline for Interactive World Simulators}
\label{sec:data}

\begin{figure}[t]
    \centering
    \includegraphics[width=\linewidth]{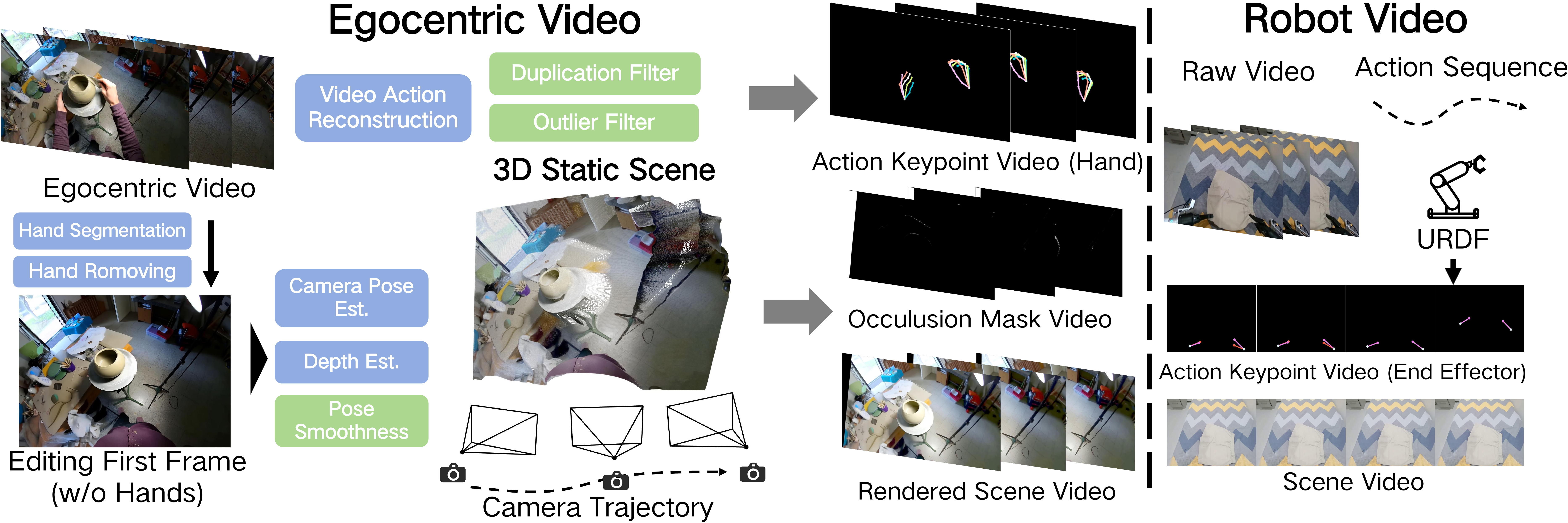}
    \caption{Overview of our scalable data processing pipeline. For both human egocentric videos and robotic manipulation videos, we automate the extraction of aligned triplets: static 3D scene point clouds, precise camera trajectories, and dynamic interaction sequences represented as spatial action keypoints.}
    \label{fig:data_pipeline}
\end{figure}

Training EgoSim requires well-aligned quadruplets: a static 3D scene, camera trajectory, embodiment-action sequence, and the corresponding interaction video. We design a fully automated pipeline (Sec.~\ref{sec:auto_pipeline}) that extracts such quadruplets from large-scale monocular egocentric and robotic videos, and further introduce EgoCap (Sec.~\ref{sec:egocap}), an intrinsics-free capture pipeline for low-cost real-world data acquisition with uncalibrated smartphones.

\subsection{Automated Data Processing Pipeline}
\label{sec:auto_pipeline}

\textbf{3D Static Scene Initialization.} 
To acquire a clean, interaction-free 3D background, we extract the first frame of each egocentric video clip. We employ SAM3~\cite{carion2025sam3segmentconcepts} to generate masks for hand regions, which are subsequently removed using the Qwen-Image-Editing~\cite{qwen3} model. Finally, DepthAnything3~\cite{depthanything3} provides monocular depth estimation, allowing us to unproject this 2D image into 3D space to form the initial scene point cloud.

\textbf{Camera Trajectory Estimation and Static Rendering.} 
To establish spatial geometric constraints, we process the original monocular video using DepthAnything3~\cite{depthanything3} to extract per-frame camera parameters, yielding the rotation matrix, translation vector, and intrinsic matrix. By rendering the initial point cloud from the perspective of this estimated camera trajectory, we produce the rendered scene video that serves as a geometry-consistent reference for all subsequent frames.

\textbf{Interaction Action Extraction and Rendering.} 
We extract a universal, cross-embodiment action representation. For human hands, we employ HaMeR~\cite{pavlakos2024reconstructing} to estimate pose sequences. We also apply duplication filters and outlier filters to enhance the annotation quality. For robotic end-effectors, we compute precise 3D action representations using recorded joint trajectories and robot URDF files. Using the extracted camera parameters, these 3D keypoints are projected onto the 2D camera plane. This operation explicitly defines the spatial action condition.

\textbf{Dataset Construction.} For training Egosim, we use our pipeline to construct a large-scale dataset with high-quality scene-interaction pairs. More details can be found in Table 3 in our supplementary.

\subsection{EgoCap: Real-World Data Collection Pipeline}
\label{sec:egocap}

\begin{wrapfigure}{r}{0.5\linewidth}
    \centering
    \includegraphics[width=\linewidth]{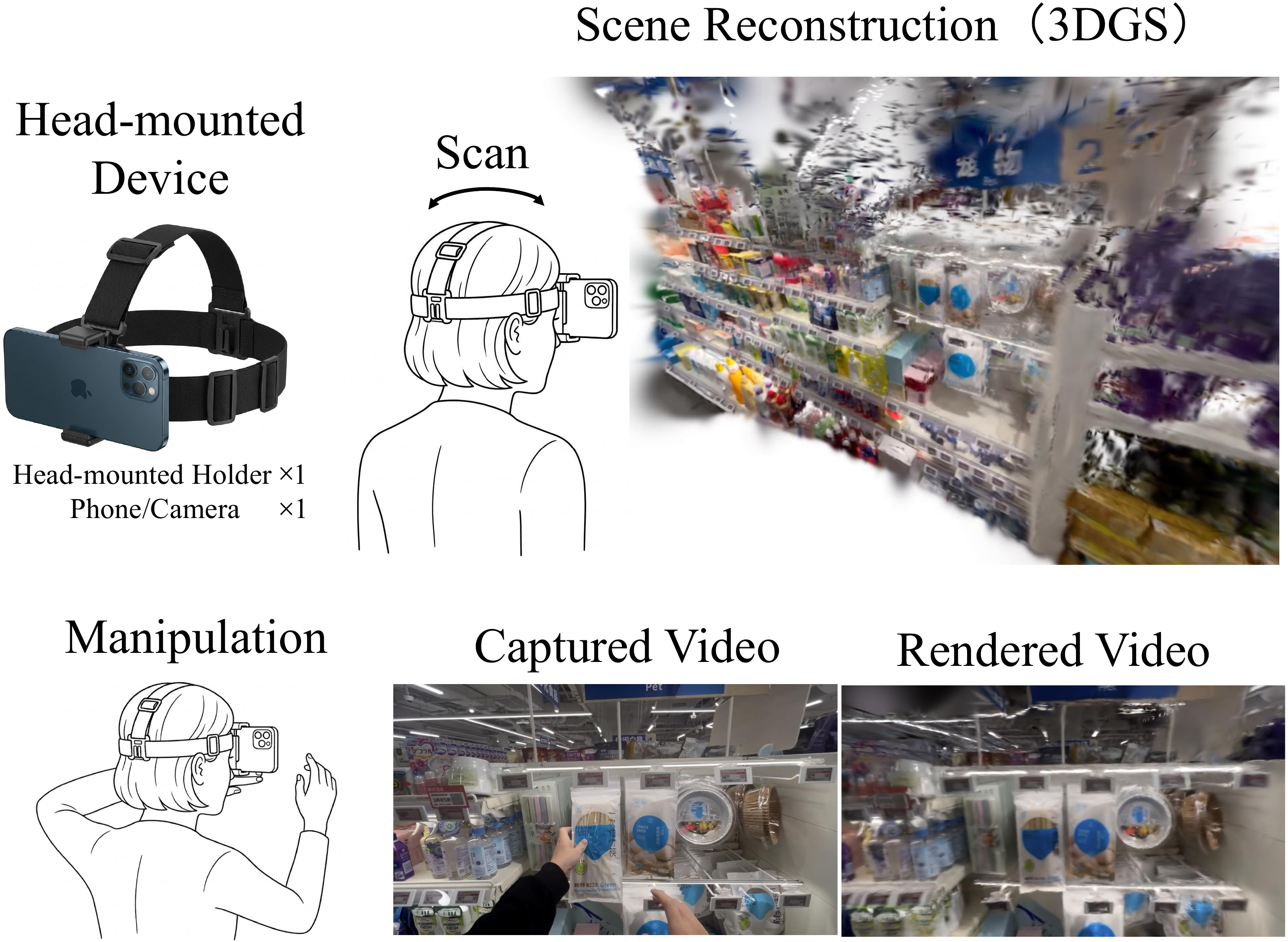}
    \caption{\textbf{The EgoCap pipeline.} An uncalibrated head-mounted smartphone first scans the scene to build a 3DGS map, then records the ego-view interaction. Relocalizing against the map yields viewpoint-aligned paired data.}
    \label{fig:egocap}
\end{wrapfigure}

Traditional collection of aligned paired data relies on rigidly calibrated multi-camera arrays, which are expensive and hard to scale. We introduce EgoCap, an intrinsic-free data collection pipeline that uses uncalibrated smartphones to produce viewpoint-aligned paired data at low cost.

\textbf{Uncalibrated Scene Scanning.} 
In the first step, a user scans the environment with an uncalibrated mobile device. Our system builds a global 3D Gaussian Splatting map using the ARTDECO~\cite{li2025artdeco} streaming reconstruction framework, with a dynamic intrinsic self-calibration module that eliminates the need for prior calibration, yielding a scale-consistent scene point cloud.

\textbf{Dynamic Interaction Recording and Relocalization.} 
The user then records the egocentric interaction with the same device. Despite unknown focal lengths, occlusions, and motion blur, a dense matching localizer continuously tracks visual features against the pre-built map to recover the 6-DoF camera trajectory.

\textbf{Continuous Trajectory Optimization and Aligned Rendering.} 
To suppress localization noise and frame discontinuities, we interpolate translations with cubic splines and rotations with SLERP, then globally smooth the trajectory using a Savitzky-Golay filter. The refined trajectory is then used to re-render the 3D scene, producing a geometry-consistent reference video aligned with the interaction viewpoint.
\section{Experiments}
\label{sec:experiment}

\subsection{Experimental Setup}
\textbf{Datasets.} We process $240\text{K}$ video clips from the EgoDex~\cite{hoque2025egodex} dataset and $160\text{K}$ video clips from the EgoVid~\cite{wang2024egovid} dataset. 
The EgoDex subset features fine-grained interactions accompanied by precise action labels. Conversely, the EgoVid subset consists of in-the-wild interaction data. 
For evaluation, we randomly select $100$ videos from each dataset to form our test set, ensuring these videos remain entirely unseen during training. All video clips used for training are uniformly processed to a length of $61$ frames at $16$ FPS.

\textbf{Training Details} 
Our backbone is initialized from Wan-2.1-Fun-14B-InP~\cite{wan2025}, generating $832 \times 480$ videos of 61 frames at 16 FPS. To incorporate action conditioning, we expand the DiT input channels to 52. We fully fine-tune the DiT for 4,000 steps using AdamW with a learning rate of $1 \times 10^{-5}$ and a per-GPU batch size of 4 across 8 NVIDIA H200 GPUs. All other components, including T5, VAE, and CLIP remain frozen. Inference uses Flow Matching with 50 steps and a CFG scale of 1.0. Further details are in the supplementary.

\textbf{Baselines.}  We compare against four methods:
(1) \textbf{Wan 2.1-14B-InP}~\cite{wan2025}, a mask-guided video inpainting model;
(2) \textbf{InterDyn}~\cite{akkerman2025interdyn}, which injects hand mask video into Stable Video Diffusion via ControlNet;
(3) \textbf{Mask2IV}~\cite{li2026mask2iv}, which conditions on both hand and object segmentation masks via latent channel-wise concatenation, enabling simultaneous modeling of hand and object dynamics;
and (4) \textbf{CosHand}~\cite{sudhakar2024coshand}, which generates each frame independently from a reference image and per-frame hand masks.
All baselines are extended to 61 frames for fair comparison.

\textbf{Evaluation Metrics.} 
We utilize a multi-dimensional metric suite to evaluate both visual quality and spatial controllability. For {Video Quality}, we measure Peak Signal-to-Noise Ratio (PSNR, $\uparrow$), Structural Similarity Index Measure (SSIM, $\uparrow$), and Learned Perceptual Image Patch Similarity (LPIPS, $\downarrow$). To evaluate spatial consistency, we adopt the metrics proposed in~\cite{wang2026hand2worldautoregressiveegocentricinteraction}.
Specifically, Depth-ERR ($\downarrow$) is utilized to measure 3D structural consistency by calculating the error between the depth maps of the generated video and the ground truth.
Cam-ERR ($\downarrow$) is employed to evaluate viewpoint consistency by computing the discrepancy in per-pixel Plücker ray embeddings.

\textbf{Evaluation Settings.} We evaluate EgoSim under two settings. In the \textbf{Normal} (single-clip) setting, the model receives a ground-truth first frame together with the 3D state rendered from the ground-truth point cloud, and generates a single 61-frame video clip. All baselines are evaluated under this setting. In the \textbf{Continuous Generation} setting, only the first ground-truth frame of the entire task is provided. The model first generates a 61-frame clip, after which the Interaction-aware State Updating module (Sec.~\ref{sec:memory}) reconstructs and updates the scene state from the generated observation; the updated state is then rendered as the spatial condition for generating the second clip, yielding a 121-frame sequence in total. We randomly select 20 full task segments from EgoDex for this evaluation.

\subsection{Qualitative Results}
\label{sec:qualitative}

\begin{figure*}[t]
    \centering
    \includegraphics[width=\textwidth]{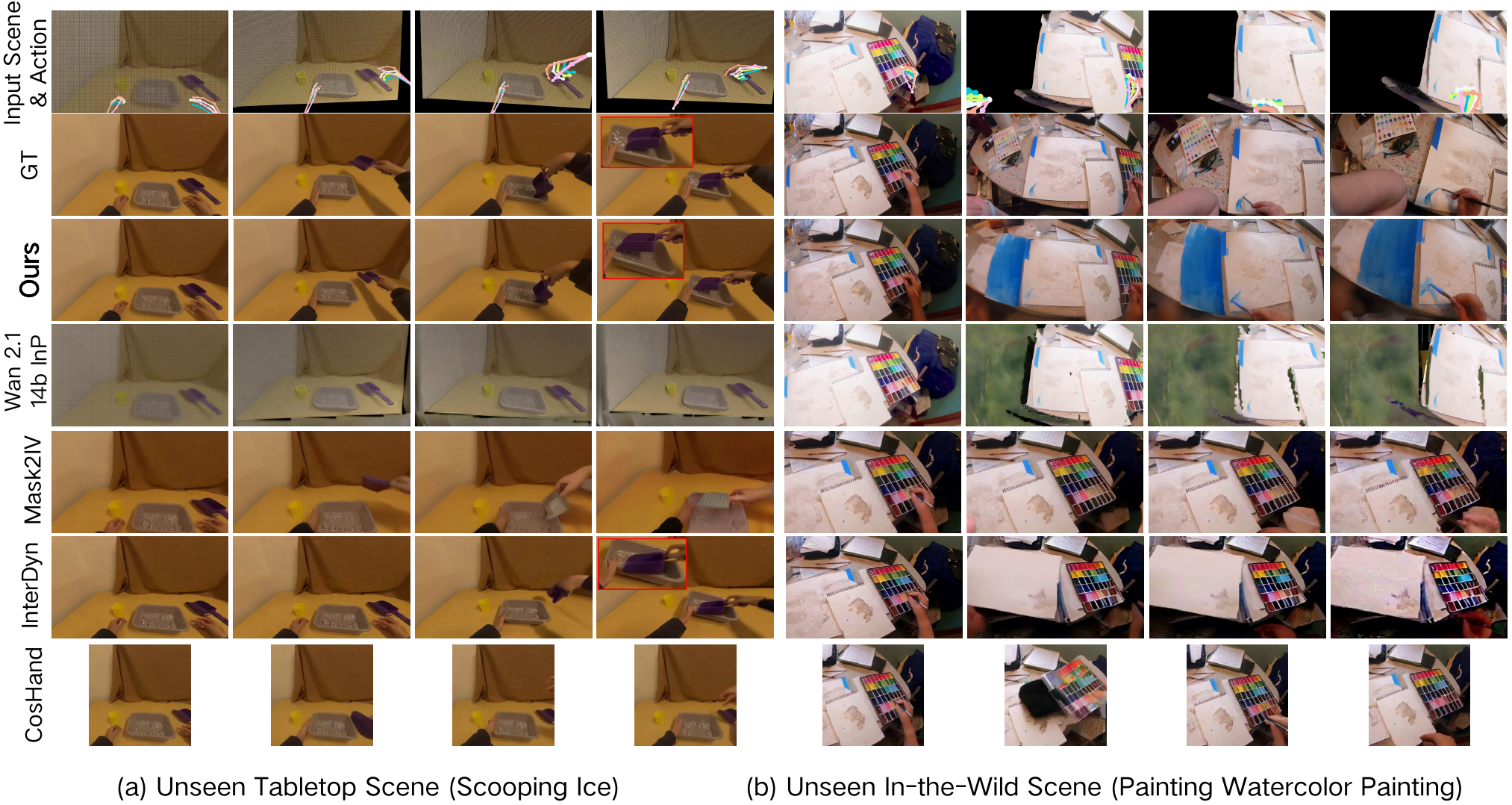}
    \caption{Qualitative comparison of EgoSim against state-of-the-art baselines on unseen test environments. (a) Inference on Egodex. (b) Inference on Egovid.}
    \label{fig:main_qualitative}
\end{figure*}

\textbf{Normal Settings.} Figure~\ref{fig:main_qualitative} provides a detailed comparison of EgoSim against the baseline methods on two unseen test scenarios.
In the \textit{scooping ice} task (Figure~\ref{fig:main_qualitative}a), EgoSim accurately simulates physical state changes and complex depth relationships, whereas InterDyn suffers from severe depth ambiguity and spatial misalignment. Mask2IV struggles to preserve the identity of hands and manipulated objects, leading to distorted interactions. CosHand, as an image generation method, produces plausible single frames but lacks temporal coherence across the sequence. In the in-the-wild \textit{watercolor painting} task (Figure~\ref{fig:main_qualitative}b), EgoSim synthesizes realistic interaction semantics and seamlessly hallucinates unobserved background regions into a coherent visual space, while Wan2.1-14B-InP acts as an identity mapping outside the mask, merely reconstructing incomplete point cloud inputs without capturing hand-object dynamics. Notably, none of the baselines can faithfully follow egocentric head motion, whereas EgoSim maintains consistent viewpoint transitions throughout the simulation.

\begin{table*}[t]
\centering
\caption{Quantitative comparison of EgoSim against baselines on the EgoDex and EgoVid test sets.}
\label{tab:quantitative_results}
\resizebox{\textwidth}{!}{%
\begin{tabular}{lcccccccccc}
\toprule
\multirow{2}{*}{\textbf{Method}} & \multicolumn{5}{c}{\textbf{EgoDex (Tabletop Scene)}} & \multicolumn{5}{c}{\textbf{EgoVid (In-the-wild)}} \\
\cmidrule(lr){2-6} \cmidrule(lr){7-11}
 & \textbf{PSNR}~$\uparrow$ & \textbf{SSIM}~$\uparrow$ & \textbf{LPIPS}~$\downarrow$ & \textbf{Depth-ERR}~$\downarrow$ & \textbf{Cam-ERR}~$\downarrow$ & \textbf{PSNR}~$\uparrow$ & \textbf{SSIM}~$\uparrow$ & \textbf{LPIPS}~$\downarrow$ & \textbf{Depth-ERR}~$\downarrow$ & \textbf{Cam-ERR}~$\downarrow$ \\
\midrule
Wan-2.1-14B-InP & 17.998 & 0.447 & 0.708 & 42.335 & 0.0300 & 11.754 & 0.430 & 0.503 & 34.470 & 0.0174 \\
Mask2IV & 20.622 & 0.814 & 0.299 & 38.339 & 0.0181 &12.311
& 0.414
& 0.571
& 34.413
& 0.0175 \\
CosHand & 17.119
& 0.776
& 0.386
& 56.524
& 0.0499 & 11.805
& 0.408
& 0.600
& 39.373
& 0.0516 \\
InterDyn & 22.250 & 0.830 & 0.255 & 44.345 & 0.0226 & 14.612 & 0.466 & 0.484 & 38.180 & 0.0308 \\
\midrule
\textbf{EgoSim (Ours)} & \textbf{25.056} & \textbf{0.896} & \textbf{0.170} & \textbf{8.888} & \textbf{0.0013} & \textbf{16.684} & \textbf{0.509} & \textbf{0.421} & \textbf{19.260} & \textbf{0.0105} \\
\bottomrule
\end{tabular}%
}
\end{table*}

\begin{table}[t]
\centering
\tiny 
\caption{Quantitative comparison of continuous generation on EgoDex.}
\label{tab:ablation_continuous}
\begin{tabular}{lccccc}
\toprule
\textbf{Model} & \textbf{PSNR}~$\uparrow$ & \textbf{SSIM}~$\uparrow$ & \textbf{LPIPS}~$\downarrow$ & \textbf{Depth-ERR}~$\downarrow$ & \textbf{Cam-ERR}~$\downarrow$ \\
\midrule
EgoSim  & \textbf{25.056} & \textbf{0.896} & \textbf{0.170} & \textbf{8.888} & \textbf{0.0013} \\
EgoSim (Continuous) & \underline{19.165} & \underline{0.835} & \underline{0.220} & \underline{10.943} & \underline{0.0017} \\
\bottomrule
\end{tabular}
\end{table}

\begin{figure}[t]
\centering
\begin{minipage}[t]{0.48\columnwidth}
\centering
\includegraphics[width=\linewidth]{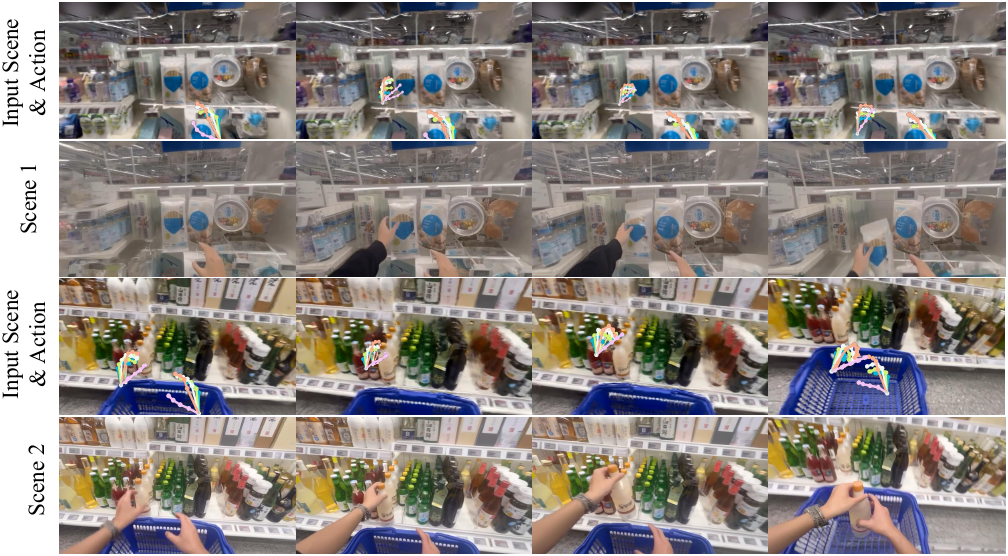}
\caption{Real-world result of EgoSim across diverse in-the-wild scenes.}
\label{fig:realworld}
\end{minipage}\hfill
\begin{minipage}[t]{0.48\columnwidth}
\centering
\includegraphics[width=\linewidth]{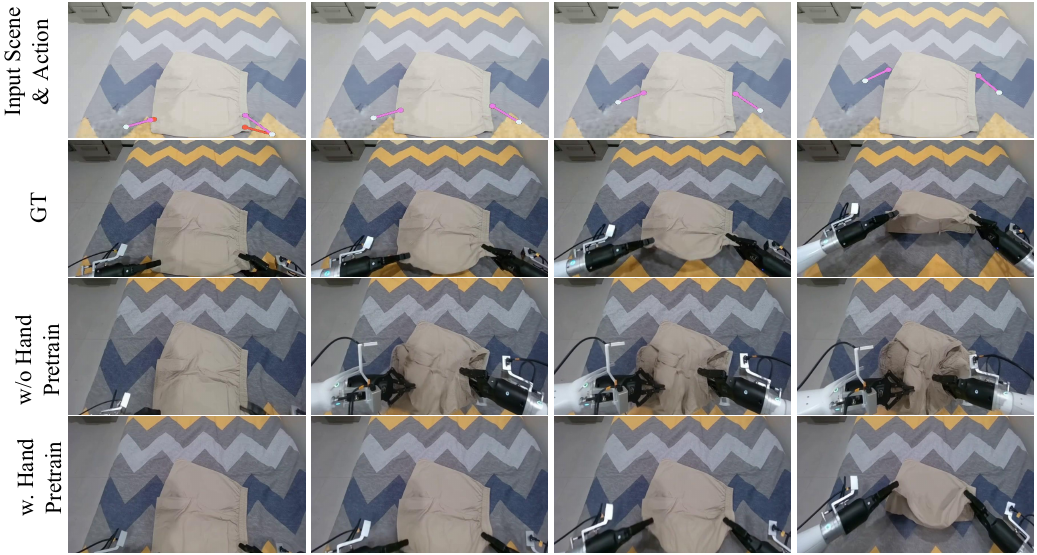}
\caption{Cross-embodiment simulation on AgiBot. \textit{w/ pretrain} model simulates complex physical dynamics accurately.}
\label{fig:cross_embodiment}
\end{minipage}
\end{figure}

Additionally, to validate the real-world data collection pipeline EgoCap, we capture 50 clips in a supermarket environment, focusing on shelf interactions, allocating 30 for training and 20 for testing. The model is initialized from our main checkpoint and fine-tuned for only 50 steps. As shown in Figure~\ref{fig:realworld}, even with this minimal fine-tuning data, EgoSim successfully synthesizes physically plausible hand-object interactions and consistent backgrounds on unseen test scenes, demonstrating efficient adaptation to novel real-world environments.

\subsection{Quantitative Comparison}

\textbf{Normal Settings.} Table~\ref{tab:quantitative_results} presents our quantitative evaluation under the single-clip setting across both the EgoDex and EgoVid test sets. The results confirm the substantial superiority of EgoSim in both visual quality and spatial controllability.

In terms of video quality, EgoSim significantly outperforms all baselines on both datasets. On the tabletop scenes of the EgoDex dataset, EgoSim achieves a PSNR of 25.056 and an SSIM of 0.896, demonstrating a substantial improvement over the mask-based InterDyn baseline. A similar trend is observed on the challenging, in-the-wild EgoVid dataset, where our method maintains the highest visual fidelity despite the complex background dynamics.

Crucially, in the spatial controllability metrics, EgoSim also outperforms the baselines. Conventional architectures like InterDyn frequently fail to decouple camera motion from hand motion, leading to severe background drifting and phantom occlusions, as reflected by their high Depth-ERR and Cam-ERR scores. By explicitly rendering point clouds along the true camera trajectory, EgoSim provides the generator with an unambiguous spatial anchor. This results in an overwhelming reduction in spatial errors. For instance, on the EgoDex dataset, our Depth-ERR is reduced to 8.888 compared to InterDyn with 44.345, and Cam-ERR is reduced by over an order of magnitude. These results unequivocally demonstrate that our spatial conditioning effectively anchors the 3D space, making EgoSim a highly spatially consistent, physically grounded simulator.

\textbf{Continuous Generation.} Table~\ref{tab:ablation_continuous} reports the results under the continuous generation setting on EgoDex, where each 121-frame sequence is generated as two consecutive clips with an intermediate 3D state update.

As presented in Table~\ref{tab:ablation_continuous}, both video quality and spatial consistency remain robust. {EgoSim} (Continuous) achieves a PSNR of $19.165$, an SSIM of $0.835$, and an LPIPS of $0.220$. While there is a marginal drop compared to the single-clip setting, as Depth-ERR increases from $8.888$ to $10.943$ and Cam-ERR from $0.0013$ to $0.0017$, this is primarily due to cumulative error from simulated artifacts and inherent noise within the updated states.

\begin{table*}[t]
\begin{minipage}[t]{0.55\textwidth}
\centering
\caption{Ablation study on EgoDex.}
\label{tab:ablation}
\resizebox{\linewidth}{!}{%
\begin{tabular}{lccccc}
\toprule
\textbf{Model} & \textbf{PSNR}~$\uparrow$ & \textbf{SSIM}~$\uparrow$ & \textbf{LPIPS}~$\downarrow$ & \textbf{Depth-ERR}~$\downarrow$ & \textbf{Cam-ERR}~$\downarrow$ \\
\midrule
w/o trajectory & 23.380 & 0.845 & 0.244 & 10.238 & 0.0015 \\
w/o mask  & 23.988 & 0.886 & 0.186 & 14.124 & 0.0022 \\
\textbf{EgoSim (Ours)} & \textbf{25.056} & \textbf{0.896} & \textbf{0.170} & \textbf{8.888} & \textbf{0.0013} \\
\bottomrule
\end{tabular}%
}
\end{minipage}
\hfill
\begin{minipage}[t]{0.45\textwidth}
\centering
\caption{Cross-embodiment eval.}
\label{tab:agibot}
\resizebox{\linewidth}{!}{%
\begin{tabular}{lccc}
\toprule
\textbf{Setting} & \textbf{PSNR}~$\uparrow$ & \textbf{SSIM}~$\uparrow$  & \textbf{LPIPS}~$\downarrow$  \\
\midrule
w/o hand pretrain & 16.36 & 0.69  & 0.31\\
w/ hand pretrain & \textbf{18.67} & \textbf{0.72} & \textbf{0.28}\\
\bottomrule
\end{tabular}%
}
\end{minipage}
\end{table*}

\subsection{Ablation Study}
To rigorously validate our architectural design choices, we conduct ablation studies on the fine-grained EgoDex dataset. The quantitative results are presented in Table~\ref{tab:ablation}.

\textbf{Effect of Camera Trajectory Rendering.} 
To verify the necessity of explicitly rendering point clouds along the true camera trajectory, we evaluate the \textit{w/o trajectory} variant. By duplicating the initial static frame across all time steps, this setting ignores camera ego-motion. As shown in Table~\ref{tab:ablation}, this causes a significant drop in generation quality, with PSNR falling to 23.380. Without explicit physical constraints from the moving 3D point cloud, the model struggles to accurately capture background parallax and hallucinates changing scene geometry.

\textbf{Effect of Mask Constraints.} 
Feeding an all-black mask (\textit{w/o mask}) still yields a high PSNR of 23.988, proving EgoSim acts as an \textit{identity function with a generative prior} that intelligently synthesizes interactions without explicit region guidance. Nonetheless, providing accurate visibility masks ({Ours}) optimally blends dynamic interactions with hallucinated unobserved regions, achieving the best PSNR of 25.056.

\begin{figure*}[t]
\centering
\includegraphics[width=\linewidth]{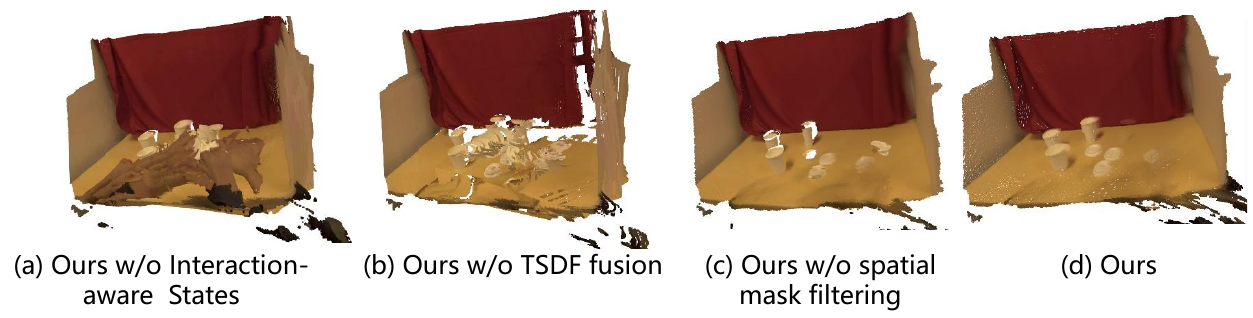}
\caption{Qualitative ablation of Interaction-aware State Updating. Our full model (d) maintains a clean, coherent 3D representation.}
\label{fig:3dmem_abla}
\end{figure*}

\textbf{Effect of Updatable Interaction-aware 3D States.} 
We validate this mechanism and ablate its key components by visualizing the post-interaction point clouds from a novel \textit{third-person perspective} (Figure~\ref{fig:3dmem_abla}). This explicitly proves that the physical displacement of objects is permanently registered in 3D space. 
Without interactive object filtering (Figure~\ref{fig:3dmem_abla}a), the static point cloud becomes corrupted by dynamic elements like hands, leading to ghosting artifacts. 
Removing TSDF fusion (Figure~\ref{fig:3dmem_abla}b) produces fragmented and noisy surface reconstructions that fail to provide a reliable geometric anchor. 
Similarly, omitting spatial mask filtering (Figure~\ref{fig:3dmem_abla}c) introduces floating artifacts and erroneous projections during dynamic interactions. 
By integrating all these components, our full pipeline (Figure~\ref{fig:3dmem_abla}d) maintains a clean, coherent 3D representation.

\subsection{Robotic Simulation with Egocentric Pretraining}
We further deploy EgoSim as a visual dynamics model for bimanual robotic manipulation using the AgiBot-World~\cite{bu2025agibot} dataset. Since the dataset is captured with a static camera, we erase the robotic arms from the initial frame via image editing to establish a clean background, and construct the static conditioning video by duplicating this frame. For the action condition, we directly project the robotic end-effector kinematics into the 3D keypoint space. We randomly sample $50\text{k}$ interaction clips for training and reserve $150$ non-overlapping clips for evaluation. To ensure a controlled comparison, both settings use the same total number of training steps: \textit{w/o hand pretrain} trains on robot data for 1,400 steps, while \textit{w/ hand pretrain} first pretrains on egocentric human hand data for 1,200 steps and then finetunes on robot data for 200 steps. 

As shown in Table~\ref{tab:agibot}, pretraining on large-scale egocentric human hand data before finetuning on AgiBot data (\textit{w/ human pretrain}) substantially outperforms robot-only training (\textit{w/o human pretrain}) across all metrics, boosting PSNR from 16.36 to 18.67, improving SSIM from 0.69 to 0.72, and reducing LPIPS from 0.31 to 0.28. Indicating better spatial and object consistency in the generated robot videos.

\section{Conclusion}
\label{sec:conclusion} 

In this work, we introduce EgoSim, a closed-loop egocentric world simulator that bridges the gap between passive video generation and active embodied participation. By anchoring visual generation to an underlying 3D point cloud and explicitly maintaining an Updatable 3D Memory, EgoSim achieves faithful spatial conditioning and long-horizon state persistence, eliminating the structural drift common in prior world models. To overcome the critical bottleneck of acquiring perfectly aligned static-dynamic paired data, we develop a scalable data construction framework that extracts aligned triplets from web-scale videos, complemented by our EgoCap tool for low-cost, real-world capture. Extensive experiments demonstrate that EgoSim excels in complex, multi-object interactions and successfully transfers to bimanual robotic manipulations. 

\noindent\textbf{Limitations:} While our pipeline successfully leverages in-the-wild data, monocular depth and camera pose estimations can occasionally fail in heavily occluded or highly dynamic environments, leading to imperfect point cloud initializations. Future work will explore integrating robust multi-view priors and physics-based contact constraints to further enhance simulation fidelity. 

\section*{Acknowledgments}

This work was supported by National Natural Science Foundation of China (No. 62595774, 72192821, 62302297, 62472282), the Fundamental Research Funds for the Central Universities (project number: YG2023QNA35), YuCaiKe [2023] Project Number: 231111310300.


%
%
\bibliographystyle{splncs04}
\bibliography{main}

\clearpage

\vspace{2cm}
\begin{center}
\Large\textbf{Supplementary Material}
\end{center}
\vspace{0.5cm}
\setcounter{section}{0}

This supplementary material provides comprehensive technical details, extended experimental evaluations, and additional results to support the main paper.

\begin{enumerate}
    \item \textbf{More Comparison Results} (Section~\ref{Sec:supp_more_quan_comparison}): We provide additional qualitative and quantitative comparisons.

    \item \textbf{Applications} (Section~\ref{Sec:application}): We provide results on downstream applications.

    \item \textbf{More Details of Interaction-aware State Updating} (Section~\ref{Sec:Implementation Updatable Interaction-aware States}): This section details the multi-stage pipeline for trajectory estimation, hierarchical interaction tracking, and multi-frame point cloud fusion.
    
    \item \textbf{Data Pipeline Details} (Section~\ref{sec: Dataset Details}): This section offers an in-depth description covering data annotation and cleaning pipelines, the development of the unified skeleton representation, and a comparative analysis with existing egocentric benchmarks.
    
    \item \textbf{EgoCap Details} (Section~\ref{sec:supp_egocap}): This section elaborates on further technical specifics regarding the EgoCap Pipeline.
    
    \item \textbf{More Experiment Details} (Section~\ref{sec:more Experiment Details}): This section provides training configurations, hardware specifications, and baseline implementation protocols as concluding content.
    
    \item \textbf{Additional Visualization Results} (Section~\ref{sec:supp_more_visualize}): This section presents additional visualization results. We also provide more visualizations in the HTML page in our supplementary.
\end{enumerate}
\section{More Comparison Results}
\label{Sec:supp_more_quan_comparison}

\subsection{More Quantitative Comparison}

\textbf{Comparison on TRUMANS Dataset~\cite{jiang2024scaling}.} To more directly evaluate whether the updated world state remains consistent with the actual 3D scene after interaction, we conduct an additional evaluation on the TRUMANS dataset, which provides ground-truth camera poses and depth for 3D-aware assessment. Specifically, we use the ground-truth camera poses and depth maps to measure the spatial consistency of generated results beyond image-level quality. As shown in Table.~\ref{tab:trumans}, We evaluate 3D consistency of EgoSim. We utilize the GT camera poses and depth from TRUMANS dataset as proper 3D metrics. EgoSim achieves superior quality and spatial consistency compared with the baseline.

\textbf{ Fine-grained metrics.} We further evaluate whether the generated hand-object interactions are physically and kinematically accurate. As shown in Table~\ref{tab:finegraind_quantitative_results}, we evaluate hand pose accuracy by calculating 2D PCK and 3D PA‑MPJPE of HaMeR‑detected keypoints from generated videos, GT keypoints from Egodex, and HaMeR keypoints from EgoVid. For object‑level pose accuracy, we report 2D and 3D AP calculated from 6D poses predicted by WildDet3D~\cite{huang2026wilddet3d} from GT and predicted videos. Due to the limitations of 6DoF detectors, we only evaluate AP on rigid‑body interactions. EgoSim achieves better 3D PA‑MPJPE compared to InterDyn with 2D hand mask priors, while maintaining better PCK in dynamic EgoVid scenes. EgoSim also outperforms the baseline across all metrics for 6 DoF accuracy.

\begin{table}[t]
\centering
\small
\begin{minipage}[t]{0.39\textwidth}
\centering
\caption{Quantitative comparison on the TRUMANS dataset.}
\label{tab:trumans}
\resizebox{\linewidth}{!}{
\begin{tabular}{lccccc}
\toprule
\textbf{Model} & \textbf{PSNR}~$\uparrow$ & \textbf{SSIM}~$\uparrow$ & \textbf{LPIPS}~$\downarrow$ & \textbf{Depth-err$_{GT}$}~$\downarrow$ & \textbf{Cam-err$_{GT}$}~$\downarrow$ \\
\midrule
InterDyn & 16.541 & 0.564 & 0.443 & 57.962 & 0.2140 \\
EgoSim & \textbf{19.599} & \textbf{0.657} & \textbf{0.406} & \textbf{16.084} & \textbf{0.0945} \\
\bottomrule
\end{tabular}
}
\end{minipage}
\hfill
\begin{minipage}[t]{0.58\textwidth}
\centering
\caption{Fine-grained interaction metrics of EgoSim.}
\label{tab:finegraind_quantitative_results}
\resizebox{\linewidth}{!}{
\begin{tabular}{lcccc}
\toprule
\textbf{Method} 
& \multicolumn{2}{c}{\textbf{EgoDex (Tabletop Scene)}} 
& \multicolumn{2}{c}{\textbf{EgoVid (In-the-wild)}} \\
\cmidrule(lr){2-3} \cmidrule(lr){4-5}
\textit{GT / HaMeR} 
& PCK@20 $\uparrow$ & PA-MPJPE $\downarrow$ 
& PCK@20 $\uparrow$ & PA-MPJPE $\downarrow$ \\
\midrule
InterDyn 
& \textbf{0.6481} & 22.82 
& 0.5962 & 23.06 \\
\textbf{EgoSim (Ours)} 
& 0.5141 & \textbf{16.81} 
& \textbf{0.6122} & \textbf{21.31} \\
\midrule
\textit{WildDet3D} 
& 2DAP$_{[.50, .95]}$ $\uparrow$ & 3DAP$_{[.05, .50]}$ $\uparrow$ 
& 2DAP$_{[.50, .95]}$ $\uparrow$ & 3DAP$_{[.05, .50]}$ $\uparrow$ \\
\cmidrule(lr){2-3} \cmidrule(lr){4-5}
InterDyn 
& 6.53 & 7.33 
& 4.99 & 3.95 \\
\textbf{EgoSim (Ours)} 
& \textbf{21.98} & \textbf{15.36} 
& \textbf{8.84} & \textbf{6.57} \\
\bottomrule
\end{tabular}
}
\end{minipage}
\end{table}

\subsection{More Qualitative Comparison}
\label{Sec:supp_more_comparison}
\begin{figure*}[t]
    \centering
    \includegraphics[width=\textwidth]{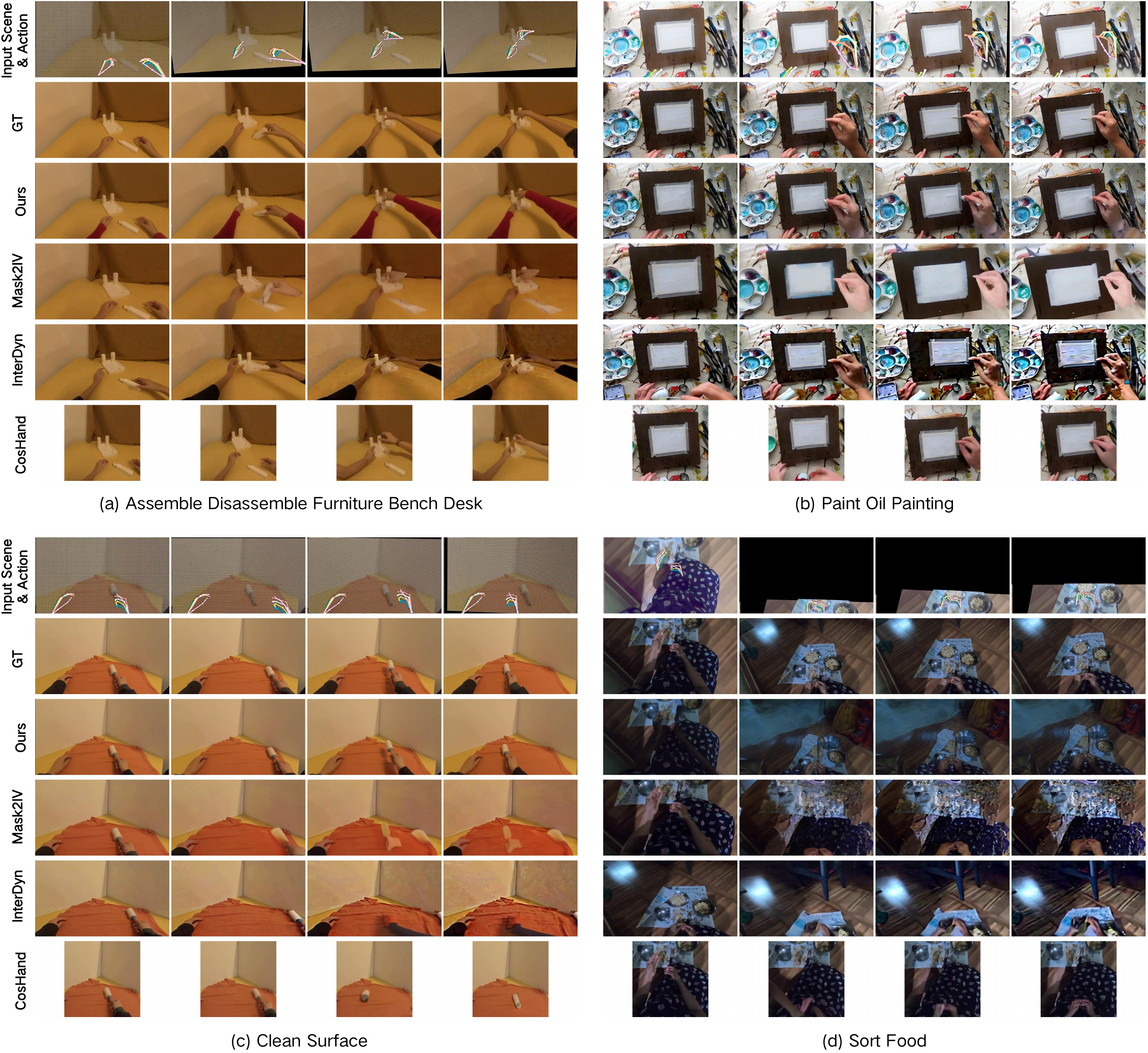}
    \caption{\textbf{More comparisons of Egosim with additional baselines.}}
    \label{fig:supp_compara_new}
\end{figure*}

We provide additional qualitative comparisons across four diverse scenarios in \cref{fig:supp_compara_new}. Compared to Mask2IV, InterDyn, and CosHand, our method consistently generates more realistic hand-object interactions and better preserves scene consistency throughout the video.

\begin{figure*}[t]
    \centering
    \includegraphics[width=\textwidth]{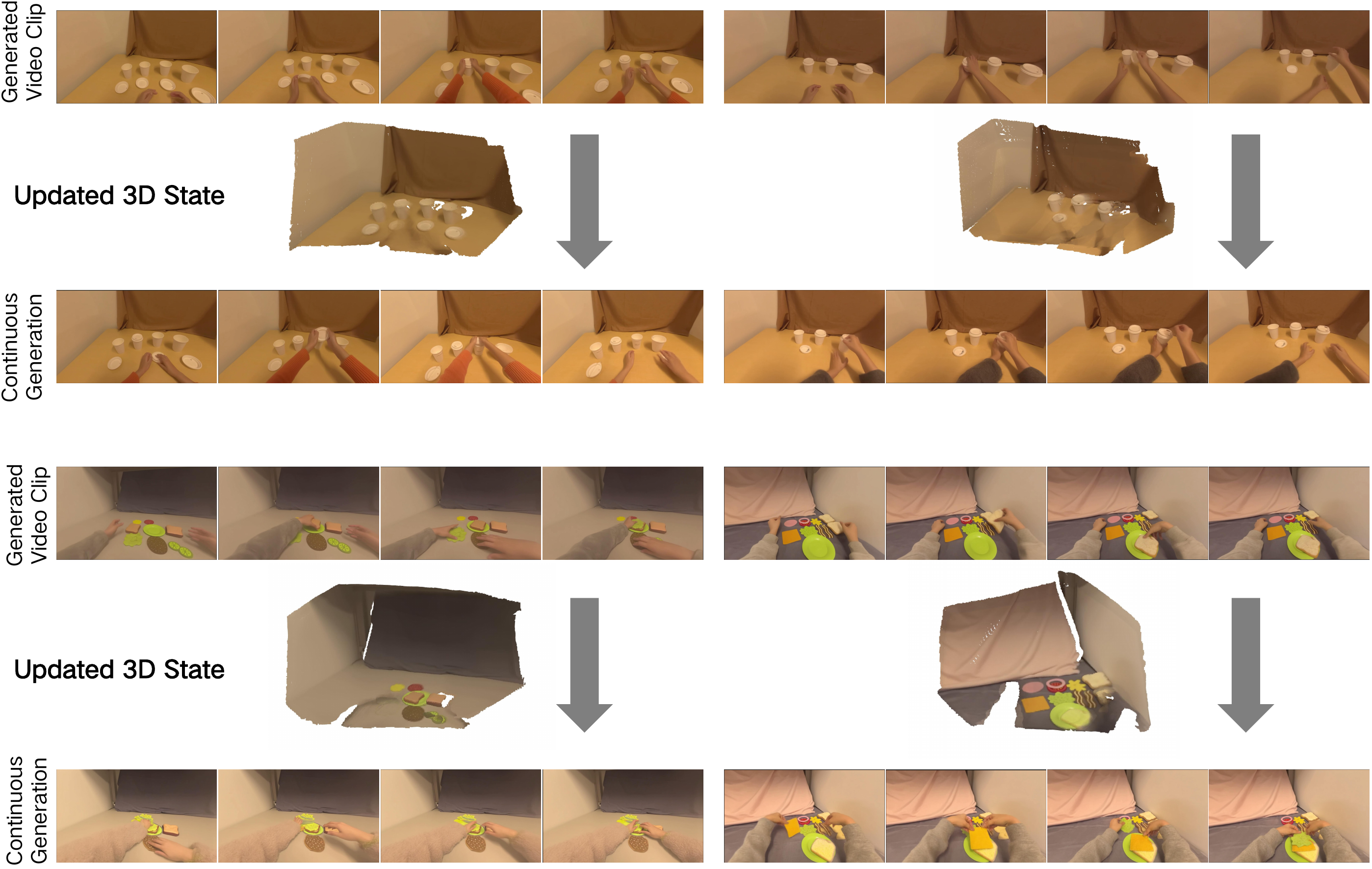}
    \caption{More simulated observations under the continuous generation setting of EgoSim, EgoSim updates dynamic interaction-aware states from the rather generated sequence and uses it as the spatial condition for simulating the following sequences.}
    \label{fig:supp_continuous}
\end{figure*}

\textbf{Continuous Generation.} Figure~\ref{fig:supp_continuous} visualizes four continuous generation sequences. Notably, the 3D states of interacted objects are correctly updated and registered within the global scene after each generation stage. For instance, in the top examples, the white cup lids are correctly closed onto the corresponding cups during the interaction process. Similarly, in the bottom sequences, breads placed on plates in the simulated observations are accurately collected and maintained in the updated states. These results demonstrate that the closed-loop state update mechanism enables {EgoSim} to propagate interaction effects across clips, producing contextually coherent long-horizon simulations.

\begin{figure}[t]
  \centering
  \includegraphics[width=\linewidth]{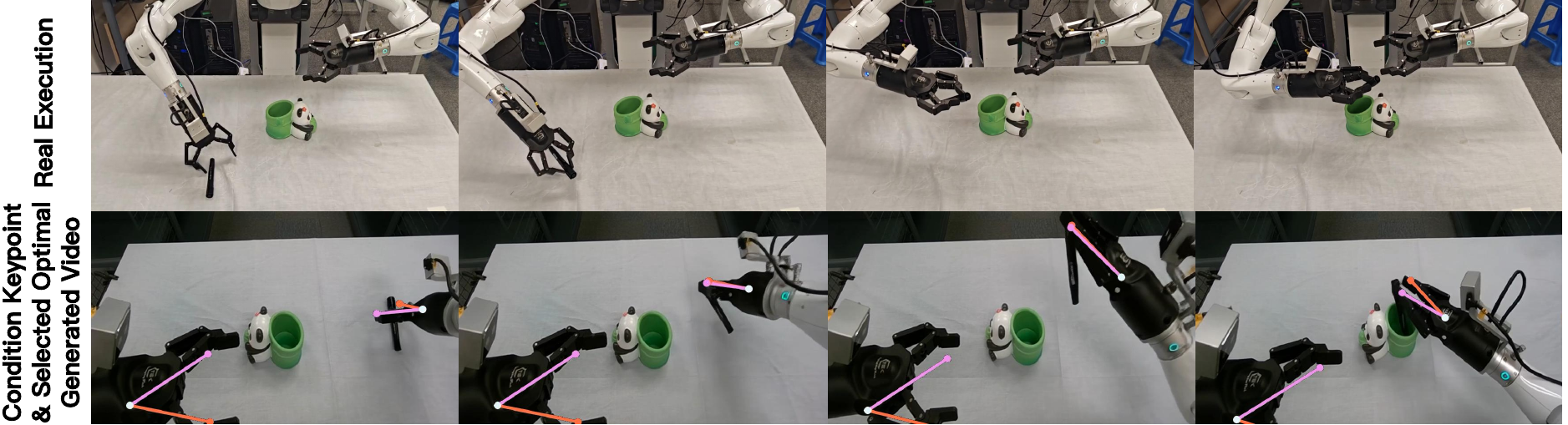}
  \caption{Robot Experiments with Model-based Planning on Task \textit{picking up a pen and placing it into a pen holder}. The pen is randomly placed for each trial.}
  \label{fig:robot}
\end{figure}

\begin{figure*}[t]
    \centering
    \includegraphics[width=\textwidth]{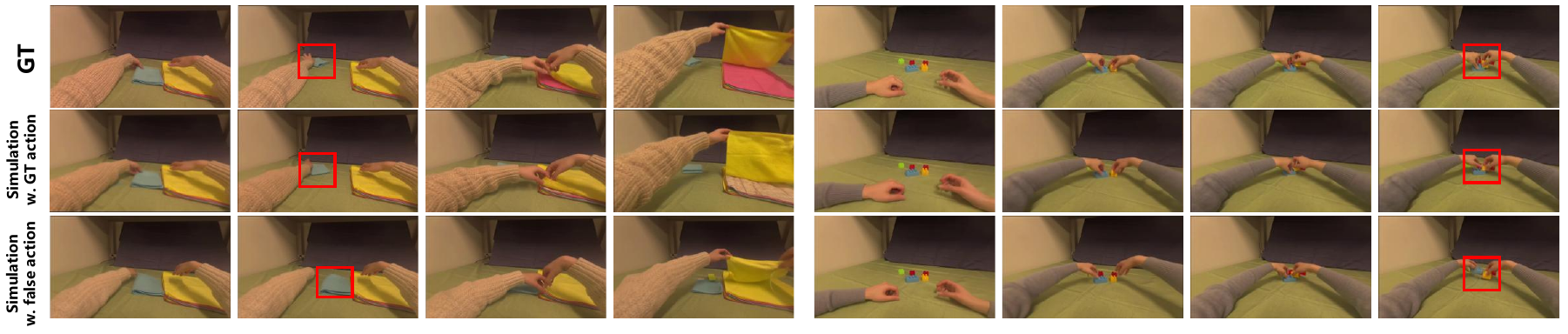}
    \caption{Egosim could generate both success and failure embodiment interactions and accordingly object dynamics following GT or incorrect action inputs.}
    \label{fig:failure_correct_case}
\end{figure*}

\section{Application}
\label{Sec:application}

\subsection{Robot Manipulation (Model-based Planning)} 
Our cross-embodiment contribution focuses on robot video generation and demonstrates potential for downstream tasks, like robot tabletop manipulation. Following the setting in DreamDojo, we further conduct real-robot experiments on the Agibot G1 (Fig~\ref{fig:robot}) to validate this potential. We collect 150 samples via teleoperation to fine-tune the VLA model Pi 0.5~\cite{intelligence2025pi_}, achieving a baseline success rate of 53.3\% (8/15) in 15 trials. To validate the effectiveness of our generated data, we adopt a model-based planning scheme. We record the current frame before each action, infer 5 times to obtain action chunks, and generate 5 corresponding videos via EgoSim and select the optimal action chunk through VLM. EgoSim planner improves the success rate from 53.3\% to 66.7\% (10/15), confirming that EgoSim-generated data can effectively boost downstream policy learning and planning for embodied AI.

\begin{figure}[t]
  \centering
  \includegraphics[width=\linewidth]{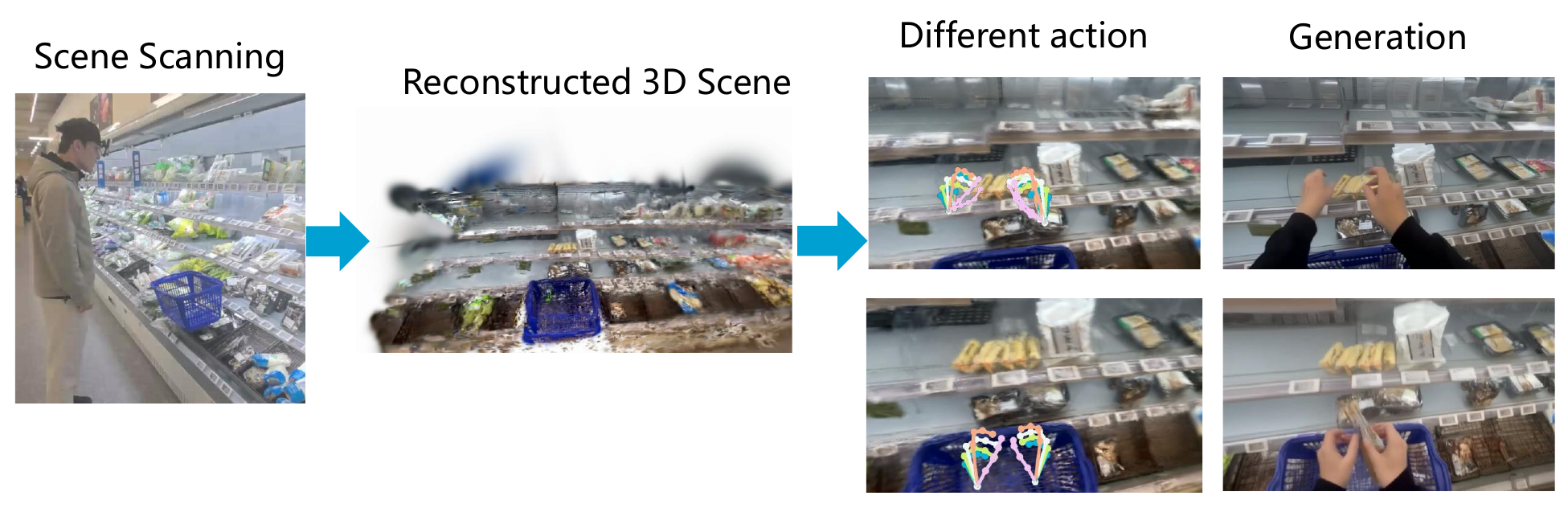}
  \caption{Egosim equipped with EgoCap could generate scalable egocentric interaction data of any novel real-life scenes.}
  \label{fig:dataengine}
\end{figure}

\begin{figure}[t]
  \centering
  \includegraphics[width=\linewidth]{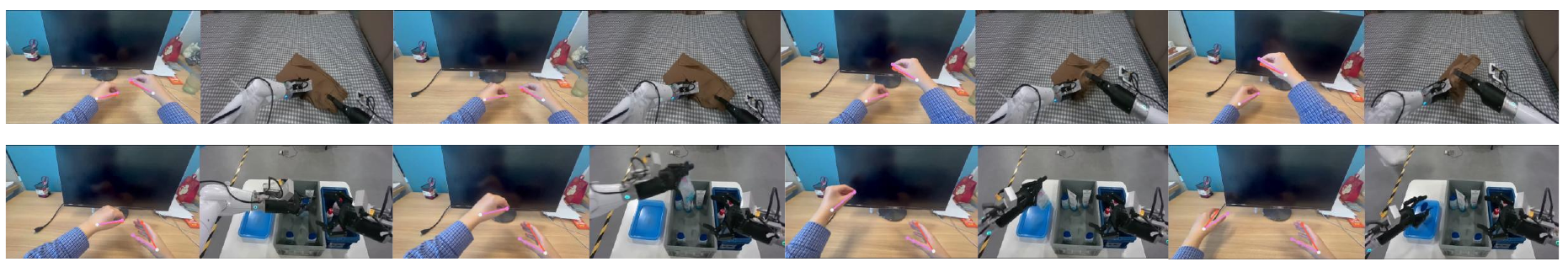}
  \caption{Egosim transfers egocentric human videos to robot manipulation videos.}
  \label{fig:human2robottransfer}
\end{figure}

\begin{figure*}[t]
    \centering
    \includegraphics[width=\textwidth]{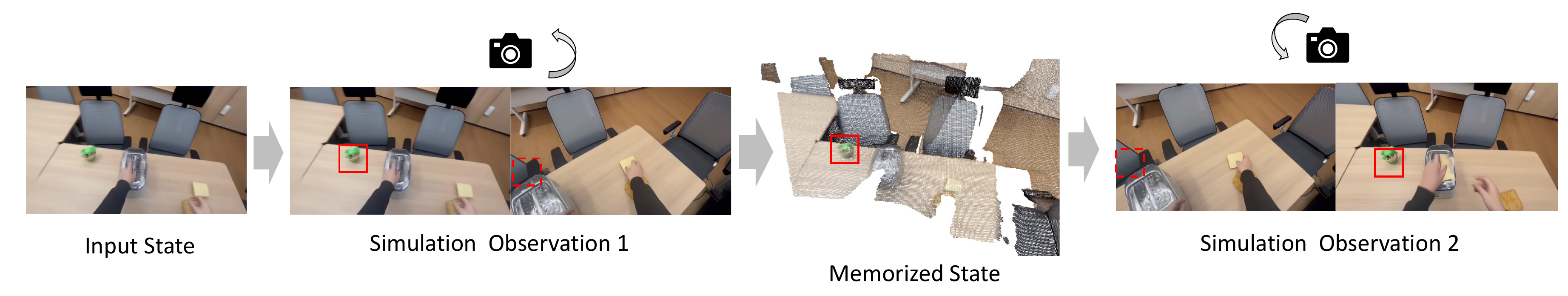}
    \caption{Egosim with updatable 3D states, cached seen environments, and enables spatial-consistency generation under large-scale dynamic view changes.}
    \label{fig:turnaround}
\end{figure*}

\subsection{Simulation both Correct and Failure Interactions} 
As an egocentric world simulator, EgoSim produces embodied interactions by strictly following the input skeleton actions. Specifically, EgoSim is capable of generating correct interaction rollouts that align with human ego data or robot teleportation ground-truths, as well as faulty interactions driven by randomized or incorrect input action trajectories. Meanwhile, EgoSim can simulate the dynamics of both embodiment and interactive objects with sound physical plausibility.
As visualized in Figure~\ref{fig:failure_correct_case}, the left case demonstrates that EgoSim yields observation sequences where the blue fabric remains unpicked when fed with perturbed incorrect action inputs, while the GT videos and observation sequence rendered with GT actions perform the same interaction. Correspondingly, the right case shows that EgoSim generates observations of improperly assembled LEGO cubes. These results validate the capability of EgoSim as a generative simulator for downstream tasks, including serving as a simulation environment for closed-loop learning of robotic policy models and the evaluation of inverse dynamic models.
Notably, EgoSim emerges the ability to generate failure interaction cases even when trained exclusively on correct human and robot interaction rollouts. We attribute this generalization ability to the diverse interaction patterns encapsulated in the large-scale in-the-wild egocentric human data~\cite{wang2024egovid, grauman2022ego4dworld3000hours}.

\subsection{Collecting Interaction Data in In-the-wild Scenes} 
EgoSim can serve as a powerful data engine for in-the-wild real-life scene simulation. Given a scene captured via a single EgoCap scan, EgoSim is able to produce scalable interactions conditioned on action inputs. As illustrated in Figure~\ref{fig:dataengine}, we first capture a 5-second video of a novel supermarket shelf scene using the RGB camera of an iPhone 13 and reconstruct a Gaussian-based scene representation via EgoCap. With only a few tuning steps on fewer than ten local manipulation clips from this scene, EgoSim can generate interactive behaviors with all static scene objects driven by diverse skeleton action inputs.

\subsection{Human to Robot Interaction Transfer}
EgoSim enables effective transfer from human demonstrations to robot manipulation videos. As shown in Figure~\ref{fig:human2robottransfer}, we first record human manipulation behaviors in real-world in-the-wild scenes and extract corresponding action trajectories via our data collection pipeline. Subsequently, we retarget human skeleton motions to the end-effector poses of Agibot-G1 and generate robotic manipulation sequences based on the transferred action inputs. The resultant robot manipulation videos exhibit behaviors consistent with human demonstrations while maintaining physical plausibility.

\subsection{Dynamic-View Simulation with Persistent 3D States} Recent world models always fail to model the out-of-view object dynamics~\cite{duan2026liveworldsimulatingoutofsightdynamics}. EgoSim could continuously generate embodiment interactions with out-of-view state dynamics. As shown in Figure~\ref{fig:turnaround}, EgoSim updates the state of green Crayon Shin-chan in an in-the-wild real-life scene, though the camera view turns around and leaves the toy, it could still successfully generate it when the view goes back, showing potential in generating controllable and spatio-temporally consistent mobile content.

\section{Interaction-aware State Updating.}
\label{Sec:Implementation Updatable Interaction-aware States}
The Interaction-aware State Updating pipeline processes raw observation sequences and outputs synchronized instance-level interaction metadata within a reconstructed 3D scene state.

\subsection{State Reconstruction}
\begin{figure*}[t]
    \centering
    \includegraphics[width=\textwidth]{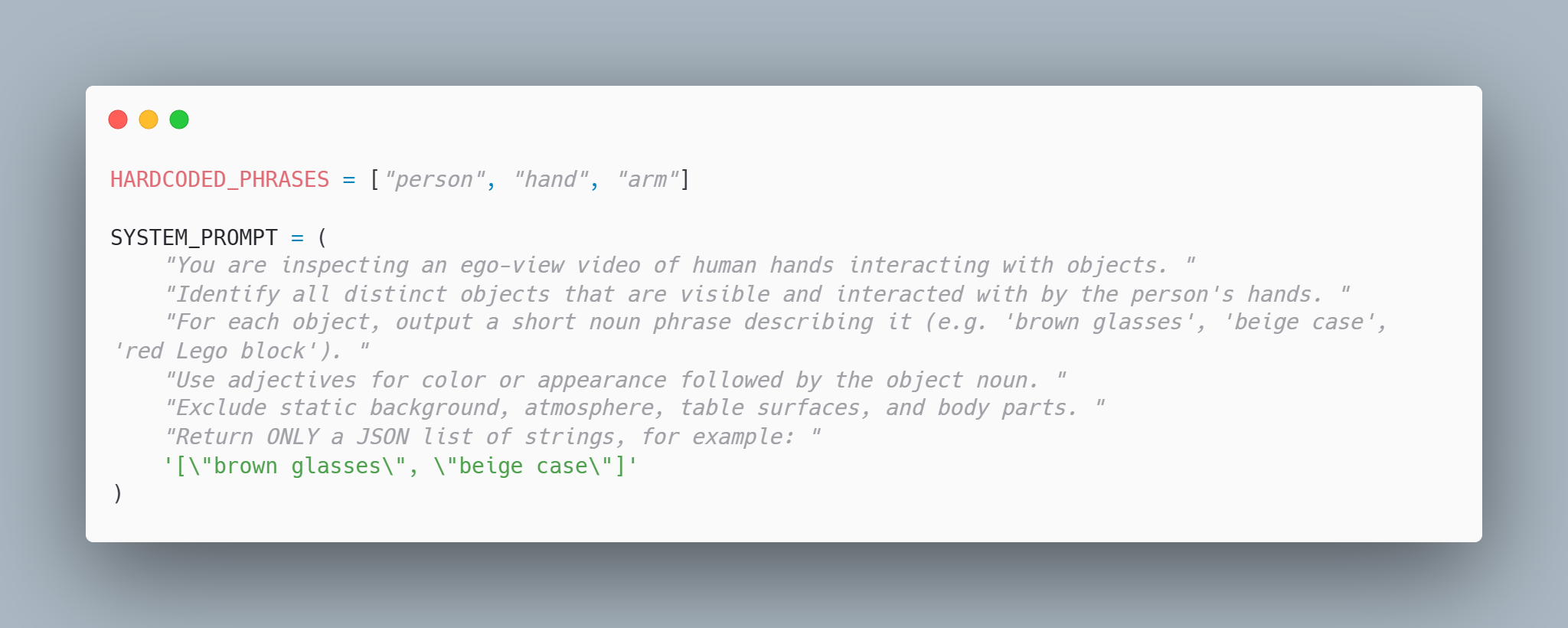}
    \caption{Prompt used for detecting interaction objects by Grounding-DINO in Updatable Interaction-aware States.}
    \label{fig:supp_groundingdino_caption}
\end{figure*}

The pipeline begins with a pre-processing stage modified from VIPE~\cite{huang2025vipe}. In this stage, the camera intrinsics are firstly estimated using the GeoCalib~\cite{veicht2024geocaliblearningsingleimagecalibration}. To establish an initial spatial understanding, we extract instance masks using SAM3~\cite{carion2025sam3segmentconcepts} guided by text phrases of objects interacted with embodiments predicted via Grounding-DINO~\cite{liu2023grounding}. The detection thresholds for bounding boxes and text are set to $0.35$ and $0.5$, respectively. The prompt of Grounding-DINO can refer to Figure~\ref{fig:supp_groundingdino_caption}, the masks of interaction objects will be tracked during the state construction along each observation sequence.

At the same time, this pipeline predicts and aligns per-frame depths and camera poses following VIPE~\cite{huang2025vipe} We employ a dual-pass {DROID-SLAM} system. During the first pass, the system tracks sparse features and selects keyframes when the tracking threshold exceeds $4.0$. These keyframes receive initial metric depth from {DepthAnything3}. In the second pass, a temporal interpolation mechanism assigns $6$-DoF camera poses and intrinsic attributes to all non-keyframe observations. Finally, a multiview depth processor aligns the predicted depth maps across different perspectives to resolve scale ambiguity.

\subsection{Interaction-aware Object State Update}
Based on the aligned geometry and continuous interaction instance tracking, we introduce a five-stage hierarchical filtering process to accurately detect and update interaction states. 

First, an initial semantic tagging is performed. Instances containing human-centric phrases, such as \textit{hands}, \textit{arms}, and  \textit{person}~\cite{jia2024orchestratingsymphonypromptdistribution}, are immediately marked as interaction references. Second, for other general objects, we evaluate their spatial proximity to the human reference. During the frame-by-frame tracking process, any object whose bounding box exhibits an IoU greater than $0.15$ with the hand bounding box is considered a potential interactive candidate.

To eliminate false positives caused by 2D projection overlap from background objects, we introduce a strict depth-based refinement. An object is confirmed as interactive only if the median depth difference between the object mask and the human reference mask is less than $0.15$m. Objects exceeding this threshold are filtered out. Furthermore, to guarantee temporal completeness, our system performs a retrospective check to update the interaction state of the last observed frame for every object. These dynamic states, including interaction status and depth distance, are serialized into a point-based updatable state to support continuous simulation.

\subsection{Incremental State Fusion}
The final phase constructs a global 3D scene state by fusing multi-frame point clouds and removing redundant geometry. To handle overlapping observations and sensor noise, we implement a Truncated Signed Distance Function (TSDF) fusion. 

The system utilizes a scalable TSDF Volume for point subsampling with a voxel size of $0.003$m, and a maximum depth truncation of $3.0$m. During integration, multi-frame observations are weighted and averaged within the voxels, which automatically handles overlap and smooths depth noise. 

We employ an incremental fusion strategy with explicit overlap detection. For the fusion of the accumulated states and constructed states from a new observation sequence, the points are projected into the camera coordinate system of the previous state. The points are integrated only if their depth discrepancy is greater than $0.05$m or if their color difference is significant. Non-interactive objects preserve their geometry from their last observed frame to maintain scene completeness. Finally, we apply a statistical outlier removal filter with $20$ neighbors and a standard deviation ratio of $2.0$ to extract a clean, global point cloud artifact.

\section{Data Pipeline Details}
\label{sec: Dataset Details}
\subsection{Introduction}
World simulators~\cite{hoque2025egodex, wang2026hand2worldautoregressiveegocentricinteraction} serve as the cognitive foundation for intelligent systems, aiming to internalize the underlying structure and causal dynamics of the physical environment to enable effective reasoning and planning. A high-fidelity world model must not only maintain a consistent representation of static scenes but also accurately predict the dynamic consequences induced by specific embodiment motion and actions, such as mobile manipulation and dexterous hand manipulations. However, transitioning from general video generation~\cite{wan2025} to faithful world simulation remains a formidable challenge. 

Developing a world model capable of such precision necessitates a fundamental shift in data requirements. It demands a new category of training data that explicitly decouples the static environment from dynamic interactions, providing ground-truth for camera trajectories, depth, and embodiment skeletons simultaneously. 

Existing egocentric datasets~\cite{kim2025dexterousworldmodels,wang2024egovid,hoque2025egodex,bu2025agibot} primarily face a trade-off between structural precision and data diversity. On one hand, synthetic datasets~\cite{kim2025dexterousworldmodels} provide perfectly aligned pairs of embodiment interactions and static scene renderings under identical trajectories, yet they suffer from a sim-to-real gap and limited diversity in interaction dynamics. On the other hand, large-scale real-world datasets collected from the web or wearable devices offer rich diversity but are often noisy, as they contain distracting body parts, inconsistent camera parameters, low-quality observations, extreme camera and embodiment movements, and especially lack precise spatial and action annotations. 

To bridge this gap, we propose a scalable and general-purpose data processing pipeline along with a novel Large-scale Heterogeneous Egocentric Manipulation Dataset. Our dataset is distinguished by the following key characteristics.

\textbf{(a) Embodiment-agnostic and Source-independent.} Unlike prior works limited to specific robots or human setups, our pipeline treats varied data from diverse web videos and open-source datasets to in-the-wild captures as a unified stream. By decoupling the interaction from specific physical embodiments, our dataset achieves true heterogeneity, enabling models to learn universal manipulation priors that transcend specific hardware or subjects.

\textbf{(b) Rich Dexterous Interactions and Mobility.} We significantly expand the scope of egocentric data beyond simple tabletop tasks with restricted camera motions. Our dataset encompasses dexterous manipulation with complex hand-object interactions, as well as large-scale camera locomotion. This variety ensures the model can perceive and predict outcomes in dynamic, non-static environments.

\textbf{(c) High-quality Multi-modal Annotations.} To provide the precise supervision required for world simulating, each clip is processed to include depth maps, camera trajectories, and action skeletons. Crucially, we provide clean static scene reconstructions by computationally removing and inpainting the ego-body and foreground occlusions, paired with high-quality automated text captions.

Our dataset serves as a robust foundation for training world models capable of generalizing to complex, real-world manipulation and locomotion tasks.
\begin{table*}[t]
\centering
\caption{\textbf{Comparison of the scalable egocentric dataset of EgoSim}  with recent open-source egocentric manipulation datasets and datasets collected by compared world simulators.} 
\label{tab:supp_dataset_campatison}
\resizebox{\textwidth}{!}{%
\begin{tabular}{lccccccccc}
\toprule
\textbf{Dataset} & {Year} & {Domain} & {Type} & {Clear scene} & {Camera} & {Depth} & {Action} & {Embodiment} & {\# Clips} \\
\midrule
Ego4D~\cite{grauman2022ego4dworld3000hours}           & 2022         & Mobile         & Real         & \xmark              & \xmark         & \xmark        & \xmark         & Human              & 3.8M              \\
Egovid~\cite{wang2024egovid}          & 2024         & Mobile         & Real          & \xmark              & Limited        & \xmark        & \xmark         & Human              & 5M                \\
Agibot-world~\cite{bu2025agibot}    & 2025         & Tabletop       & Real         & \xmark              & \cmark         & \xmark        & Joint          & Robot              & 1M                \\
Egodex~\cite{hoque2025egodex}          & 2025         & Tabletop       & Real         & \xmark              & \cmark         & \xmark        & Skeleton       & Human              & 338K              \\ \midrule
DWM~\cite{kim2025dexterousworldmodels}          & 2025         & Mobile        & Sim \& Real         & \cmark              & \cmark         & \xmark        & mesh      & Human             & 110K              \\
Hand2World~\cite{wang2026hand2worldautoregressiveegocentricinteraction}          & 2026         & Tabletop       & Real         & \cmark              & Limited         & \xmark        & mesh       & Human              & $< 8K$              \\
\midrule
\textbf{Ours}   & 2026         & Mobile         & Real          & \textbf{\cmark}     & \textbf{\cmark} & \textbf{\cmark} & \textbf{Unified Skeleton} & \textbf{Mixed} & \textbf{500K (Scalable)} \\
\bottomrule
\end{tabular}%
}
\end{table*}

\subsection{Data Annotation Pipeline}
We provide a comprehensive description of the dataset collection pipeline of EgoSim. This pipeline consists of four primary stages, including clip standardization, static scene reconstruction, action representation, and prompt generation. The codes of the data annotation pipeline will be released soon.

\textbf{Clip standardization.} We construct our training corpus by integrating diverse human and robotic manipulation datasets. We construct training clips into a uniform format of 61 frames per clip at 16 fps. For Human Datasets, we utilize EgoDex~\cite{hoque2025egodex} and EgoVid~\cite{wang2024egovid}. Specifically, for EgoDex, we directly segment the raw footage into clips. For EgoVid, we adopt their original clips and convert them to our format. For robot datasets, we incorporate a subset of the Agibot-World-Beta dataset from task indices 327 to 475, focusing on 30 daily tabletop manipulation tasks. We segment 100K high-quality clips from 25,507 provided center-camera videos. The detailed task distribution is summarized in Table~\ref{tab:task_distribution_summary}.

\textbf{Static scene reconstruction.}
To decouple dynamic interactions from the environment, we employ a mask-inpainting strategy to obtain clean and static scene backgrounds. 

(a) Egocentric body-part Masking. We first employ SAM3~\cite{carion2025sam3segmentconcepts} to detect and mask ego-body parts, like hands and arms, using the phrases \{\textit{hand, arm, person}\}, generating precise red masks. For robotic datasets where SAM3 struggles with mechanical textures, we directly inpaint and discard their arms.

(b) Embodiment Inpainting. We utilize the Qwen-Image-Edit~\cite{qwen2.5} model to remove the masked regions. For human data, we use the prompt, \textit{Discard and remove the hand and arms according to the red masks of this image, keep the other part the same}. For robotic data, we specifically address interaction dynamics with the prompt, \textit{Remove and discard the robot arms and grippers in this image, and fill the discarded area with suitable background}.

(c) Spatial Geometry Estimation. To reconstruct the 3D structure, we leverage DepthAnything3~\cite{depthanything3} to predict camera trajectories, camera intrinsics, first-frame point clouds, and per-frame depth maps. By rendering the static scene with the first-frame point clouds and predicted camera trajectories, we could obtain the videos of point-based spatial conditions for EgoSim.

(d) Unified Action Representation. A key challenge in enhancing robots with generative world simulators is the embodiment gap between human hands and robot grippers. We propose a unified skeleton representation to achieve embodiment-agnostic action modeling. For human actions, we utilize a unified 21-keypoint MANO~\cite{MANO:SIGGRAPHASIA:2017} skeleton to represent their hand actions. For EgoDex, we transform their ARKit-style 26-keypoint skeleton annotations into 21-keypoint MANO skeletons by deleting the redundant hand keypoints. For datasets lacking hand annotations, we utilize the Hamer~\cite{pavlakos2024reconstructing} model to extract hand poses, as shown in Figure~\ref{fig:supp_visualize_egovid}. For robot actions, we synthesize skeleton-based annotations using end-effector (EE) poses, joint states, and URDF of corresponding robot embodiments. To maintain consistency with human hand structures, we map the EE status to a simplified skeleton comprising the index finger and thumb, while simultaneously registering the opening state of the grippers, as shown in Figure~\ref{fig:supp_visualize_agibot}.

(d) Prompt Generation. To provide EgoSim with high-level semantic guidance, we generate descriptive prompts for each clip. For human data, we employ the Qwen3VL-8B~\cite{qwen3} model to generate detailed scene and action comprehensions, following the descriptive style established in EgoVid~\cite{wang2024egovid}. For robot data, given the structured nature of robotic tasks, we directly utilize the provided task descriptions as the ground-truth text prompts.

\subsection{Data Cleaning Pipeline}
Real-world egocentric data often contains distracting body parts, inconsistent camera parameters, and extreme jitter. We implement a robust cleaning pipeline to enhance spatial-temporal consistency.
\subsubsection{Outlier Detection and Correction}
The raw camera trajectories predicted by Depthanything3~\cite{depthanything3} sometimes exhibit sudden jumps. For each frame $i$, we evaluate its deviation from adjacent frames $i \pm 1$ based on translation ($t$) and rotation ($q$). A frame is flagged as an outlier if:
\begin{equation}
    \|t_i - t_{i \pm 1}\| > \delta_{trans} \quad \text{and} \quad \theta(q_i, q_{i \pm 1}) > \delta_{rot}
\end{equation}
where $\delta_{trans} = 0.1m$ and $\delta_{rot} = 50^\circ$. Outliers are repaired using the linear mean for translation and the normalized average of the neighborhood $[i-2, i+2]$ for rotation.

\subsubsection{Pose Kalman Smoothing}
To suppress high-frequency noise, we employ a {Pose Kalman Filter} that includes both translation and rotation smoothing.

\textbf{Translation Smoothing.} We define a 6D state vector $S_t = [x, y, z, v_x, v_y, v_z]^T$. We assume a constant velocity model for prediction and use the observed coordinates for updates. We set the process noise $Q = 1 \times 10^{-4}$ and measurement noise $R = 1 \times 10^{-3}$, ensuring the filter remains responsive to actual movement while suppressing jitter.

\textbf{Rotation Smoothing.} We maintain a 4D quaternion state $q$. The prediction step assumes $q_{pred} = q_{prev}$. After the update step using the observed quaternion, we perform re-normalization:
\begin{equation}
    q_{final} = \frac{q_{updated}}{\|q_{updated}\|}
\end{equation}
This ensures the rotation remains a valid unit quaternion throughout the sequence.

\subsubsection{Action Filtering and Refinement} 
While modern human hand estimation models~\cite{pavlakos2024reconstructing} have achieved promising performance, they frequently exhibit instability when processing unconstrained in-the-wild egocentric videos. These models are particularly susceptible to motion blur, heavy occlusion, and low-resolution observations. To address these challenges and ensure high-fidelity action labels for EgoSim, we implement a robust filtering pipeline.

\textbf{Frame-wise Spatial Filtering and De-duplication} 
We process each frame independently to resolve redundant detections. For each frame $i$ in the sequence, we derive 3D joints from MANO parameters including \textit{global\_orient}, \textit{hand\_pose}, \textit{betas}, and \textit{cam\_t}. These 3D joints are projected onto the 2D image plane to obtain 21-keypoint coordinates for computing the hand bounding box $B_i$. We utilize an Intersection-over-Union (IoU) threshold to identify duplicates and employ a Breadth-First Search (BFS) to group overlapping detections, retaining only the instance with the highest confidence score for each hand side.

\textbf{Temporal Outlier Detection} 
To ensure temporal smoothness, we identify sudden spikes in hand motion that signify tracking failures. We evaluate the dynamics between the previous ($i-1$), current ($i$), and next ($i+1$) frames by calculating the translational velocity $v_t$ in camera space through the following expression
\begin{equation}
v_t = \frac{\| \textit{cam\_t}_{curr} - \textit{cam\_t}_{prev} \|}{\Delta t}
\end{equation}
A frame is classified as a temporal outlier if it exhibits a {bidirectional jump} where both the incoming velocity and outgoing velocity exceed predefined thresholds. Such frames are discarded to prevent the model from learning physically impossible motion discontinuities.

\subsection{Dataset Statistics}
To further evaluate the superiority of our proposed pipeline, we conduct a comprehensive comparison with recently released and widely-used egocentric manipulation datasets, as summarized in Table~\ref{tab:supp_dataset_campatison}.

\textbf{Limitations of Existing Benchmarks.} Current egocentric datasets often face a trade-off between structural precision and data scalability. While foundational datasets like Ego4D~\cite{grauman2022ego4dworld3000hours} provide massive scale 3.8M clips, they lack essential 3D grounding, such as depth information and precise action labels. Conversely, recent efforts such as Agibot-world~\cite{bu2025agibot} and EgoDex~\cite{hoque2025egodex} introduce structured action representations like joint angles or skeletons. However, these datasets are typically restricted to static tabletop environments, failing to generalize to complex mobile scenarios.

\textbf{Scalability and Diversity.} In contrast, our dataset is the first to achieve both large-scale diversity and high-fidelity structural alignment in mobile egocentric settings. By leveraging an automated processing pipeline on diverse web data, we provide 500K scalable clips that encompass broad mobile domains, involving significant locomotion and varied camera ego-motion with fully aligned camera trajectories and depth maps for every clip, ensuring a consistent physical grounding.

\textbf{Bridging Human and Robot Actions.} A key innovation of our dataset is the introduction of a Unified Skeleton representation, which addresses the long-standing challenge of data heterogeneity across different embodiments. We map diverse hand and robot arm configurations from multiple sources into a standardized five-finger skeleton format, as shown in Figure~\ref{fig:supp_visualize_egovid}. To bridge the gap between human hands and robotic end-effectors, inspired by~\cite{cai2025innonscalingegocentricmanipulation}, we represent robotic grippers using the indices of the human thumb and index finger. This unified formulation allows our model to learn universal manipulation priors from a Mixed embodiment pool with potential in scaling up to more different embodiment types, as shown in Figure~\ref{fig:supp_visualize_agibot}.

\section{Detailed Implementation of EgoCap Pipeline }
\label{sec:supp_egocap}

We detail the Egocap internal reconstruct-then-relocalize mechanisms, which robustly recover precise 6-DoF camera trajectories from uncalibrated videos.

\textbf{Intrinsics-Free Scene Reconstruction.} During the initial scanning phase, the system extracts frames at a fixed rate (20 FPS) and invokes COLMAP for automatic intrinsic self-calibration, jointly estimating focal length and lens distortion without prior device profiles. Building upon the ARTDECO~\cite{li2025artdeco} framework, the pipeline incrementally processes these frames to perform joint SLAM and 3D Gaussian Splatting (3DGS) optimization. This yields a scale-consistent global 3DGS map, which is subsequently converted into a standard 3D point cloud via opacity-based filtering to remove low-confidence primitives, providing a clean static scene representation.

\textbf{Robust Camera Relocalization.} To recover the interaction trajectory under severe occlusions and motion blur, the system samples the interaction video densely (30 FPS) and employs a dual-path relocalization architecture. The primary path utilizes the ARTDECO dense matching localizer to establish frame-to-map correspondences against the pre-built 3D map, independently recalibrating intrinsics to account for unknown focal lengths. If tracking success falls below 70\% due to extreme motion, the system gracefully degrades to a fallback path. This path performs COLMAP-based offline registration and subsequently aligns the local trajectory to the unified scene coordinate system via a Sim(3) similarity transformation with outlier rejection.

\textbf{Trajectory Smoothing and Aligned Rendering.} Since unconstrained handheld capture inevitably introduces localization noise and frame discontinuities, we apply a multi-stage temporal refinement to the raw pose estimates. The system detects anomalous jump frames and applies cubic spline interpolation for translations alongside spherical linear interpolation (SLERP) for rotations, ensuring smooth spatial transitions while preserving unit quaternion constraints. A global Savitzky-Golay filter then suppresses high-frequency jitter. Based on this continuous 6-DoF trajectory, the system re-renders the initial 3DGS map to produce a geometry-consistent static background video, thereby forming the strictly aligned paired training data required by the EgoSim framework.

\section{More Experiment Details}
\label{sec:more Experiment Details}

\subsection{Training \& Inference Details.}
\textbf{Foundation Model Pre-training.} Our base video diffusion model is initialized from Wan2.1-Fun-14B-InP, generating output simulations at a resolution of $832 \times 480$ with 61 frames at 16 FPS. To accommodate hand action conditioning, we expand the Diffusion Transformer (DiT) input channels to 52. We apply full fine-tuning to the DiT for 4,000 steps, while all other components, including T5 text encoder, VAE, and CLIP image encoder, are remain frozen. We employ the AdamW optimizer with a learning rate of $1 \times 10^{-5}$. Training is conducted on a computing node equipped with 8 NVIDIA H200 GPUs, using a per-GPU batch size of 4.

\textbf{Cross-Embodiment Fine-Tuning.} For robotic manipulation evaluations on the Agibot dataset, we randomly select 50K clips from the processed 100K for training and 150 non-overlapping clips for testing. All clips are processed to 61 frames at 16 FPS. We explore two initialization settings. (1) \textit{Without hand data pre-training}, which trains directly from the expanded Wan2.1-Fun-14B-InP model, and (2) \textit{With hand data pre-training}, training from our EgoSim model pre-trained for 4,000 steps. Both variants are fine-tuned for an additional 200 steps utilizing identical hyperparameters as the main training phase.

\textbf{Real-World Fine-Tuning.} To validate our real-world data collection pipeline, EgoCap, we capture 50 clips in a supermarket environment, focusing on shelf interactions. Following the standard format of 61 frames and 16 FPS, we allocate 30 clips for training and 20 for testing. The model is initialized from our main 4,000-step checkpoint and fine-tuned for 50 steps. For this dataset, we adjust the per-GPU batch size to 1 while keeping other optimizer settings consistent.

\textbf{Inference Settings.} During the inference phase, all models across different datasets and evaluation settings employ the same generation hyperparameters. We use the Flow Matching sampling process with 50 inference steps. To preserve the generative prior and balance fidelity with action-conditioning adherence, the classifier-free guidance (CFG) scale is consistently set to $1.0$.

\textbf{Inference of Continuous Simulation.}
To enable the state updating along the dynamics of out-of-view contents and spatio-temporal consistency in continuous observation simulation, we propose and implement a novel updatable 3D interaction-aware state.

In the continuous generation setting, we formulate the updated 3D state, constructed from previously generated observation sequences, as the spatial 3D condition for simulating the subsequent sequence. We utilize the input actions $H_{k+1}$ and camera trajectories $C_{k+1}$ to render the accumulated state$S_{k}$, thereby obtaining the rendered Scene Observation $O_{k+1}$. This rendering serves as a geometric anchor, ensuring that all previously modified object states are preserved and correctly projected into the new simulation cycle.

\subsection{Baseline Details}
\label{sec:baseline_details}

\textbf{InterDyn.} InterDyn~\cite{akkerman2025interdyn} extends Stable Video Diffusion (SVD) with a ControlNet branch. Given a single conditioning image $\mathbf{I}0$ and a temporally aligned sequence of hand-object segmentation masks $\{\mathbf{M}_t\}$ as control signals, the model generates a fixed-length video clip pf 14 frames in one forward pass via score-distillation sampling. To produce a 61-frame output for fair comparison with our method, we adopt an autoregressive inference strategy. The full mask sequence is pre-divided into consecutive non-overlapping segments of equal length. At each step, the model is conditioned on the last frame of the previously generated segment as the new starting image, together with the corresponding mask sub-sequence, to generate the next segment. The resulting segments are concatenated by discarding the first frame of each subsequent segment, which is identical to the last frame of the preceding segment, yielding a seamless 61-frame video.

\textbf{Mask2IV.} Mask2IV~\cite{li2026mask2iv} is a latent diffusion model that jointly conditions on an initial frame and a mask video sequence. At each inference call, the initial frame is replicated $T$ times to construct the input video tensor $\mathbf{V} \in \mathbb{R}^{1 \times C \times T \times H \times W}$; the corresponding $T$-frame mask segment is encoded into the latent space and concatenated channel-wise with the replicated frame latent, forming a compound conditioning signal. A text prompt is additionally injected via cross-attention. The model generates $T{=}16$ new frames per call using DDIM sampling. To extend generation to 61 frames, we apply the same autoregressive chaining strategy: beginning from the first frame of the ground-truth video, we iteratively produce 16-frame segments, each time feeding the last predicted frame as the initial frame for the next segment, while the mask for segment $i$ is sliced from the full-length mask video as frames $[i \cdot T,\ (i{+}1) \cdot T)$. Four segments are generated and concatenated with duplicate boundary frames removed.

\textbf{CosHand.} CosHand~\cite{sudhakar2024controlling} generates videos in a non-autoregressive manner. Given a single reference frame from the video clip and a sequence of target hand masks specifying the desired hand position at each timestep, the model performs an independent DDIM diffusion sampling for each frame. The hand geometry is injected into the UNet via latent concatenation, while the semantic appearance of the reference frame is provided through CLIP features via cross-attention conditioning. In total, 61 frames are generated independently and arranged chronologically to form the final predicted video. Since each frame is generated in isolation without any temporal dependency on neighboring frames, the model lacks an explicit mechanism to ensure inter-frame consistency, which is one of its primary limitations compared to autoregressive video generation approaches.

\section{More Visual Results}
\label{sec:supp_more_visualize}

In this section, we provide additional qualitative results to further demonstrate the versatile simulation capabilities of EgoSim across diverse scenarios and embodiments. These results showcase our method's ability to handle dexterous manipulation, large viewpoint movements, deformable objects, and realistic object dynamics. We also provide more visualizations in the HTML page in our supplementary.

\begin{figure*}[t]
    \centering
    \includegraphics[width=\textwidth]{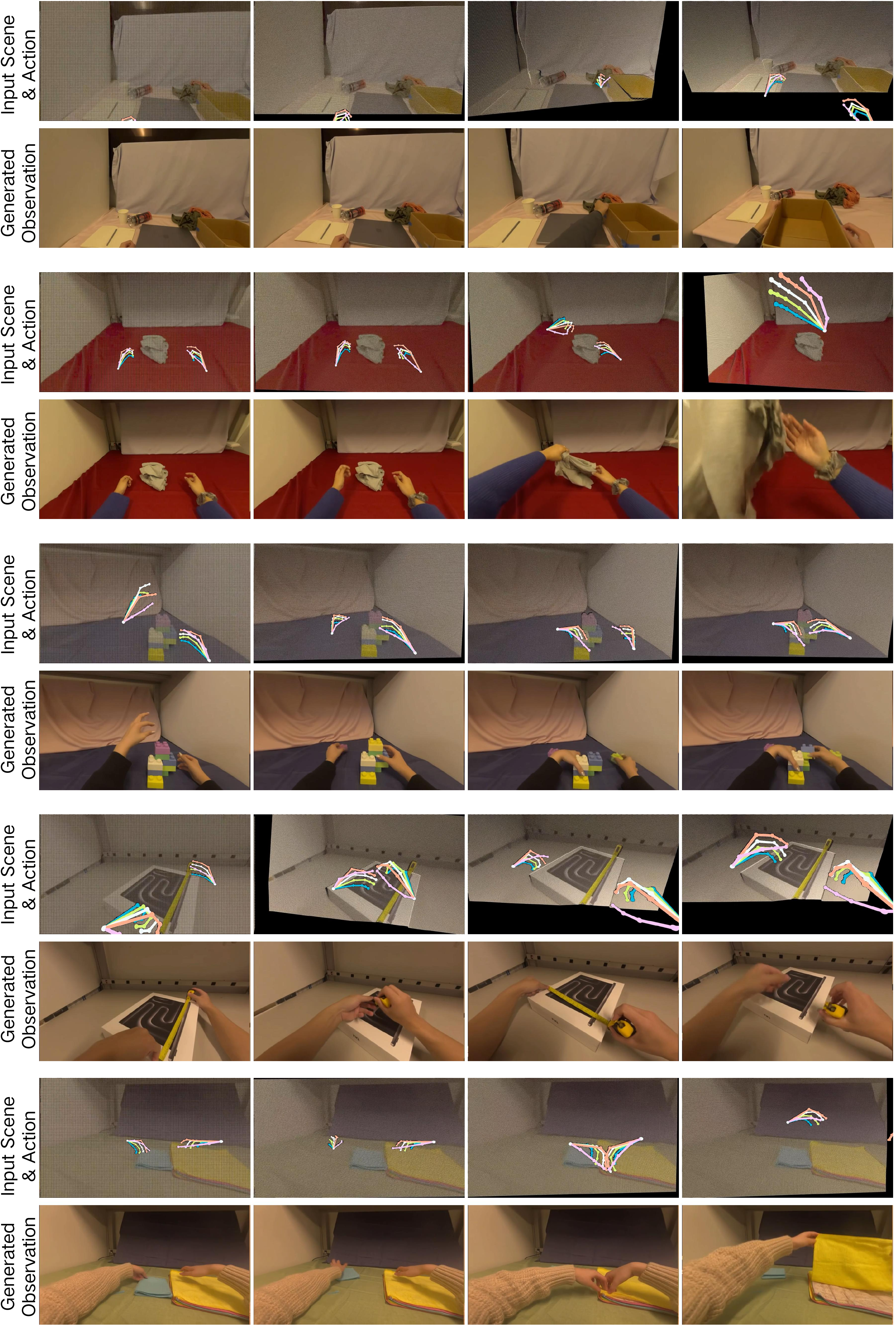}
    \caption{\textbf{Visualization} of EgoSim simulating test samples of Egodex, highlighting the ability for generating difficult dexterous manipulations over soft, small, or assembled objects.}
    \label{fig:supp_visualize_egodex}
\end{figure*}

\begin{figure*}[t]
    \centering
    \includegraphics[width=\textwidth]{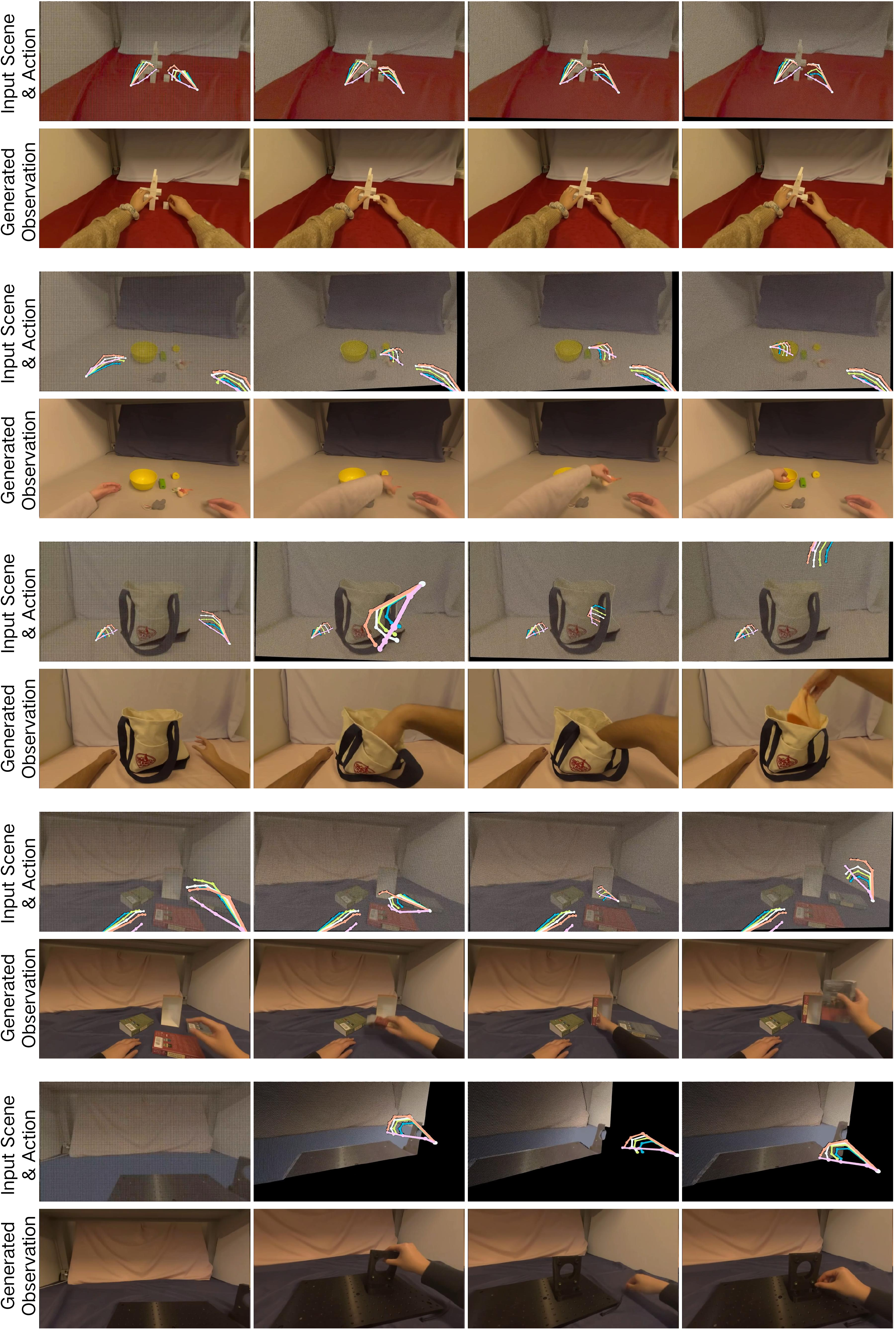}
    \caption{\textbf{Visualization} of EgoSim simulating test samples of Egodex, highlighting the ability for generating difficult dexterous manipulations over soft, small, or assembled objects.}
    \label{fig:supp_visualize_egodex2}
\end{figure*}

\begin{figure*}[t]
    \centering
    \includegraphics[width=\textwidth]{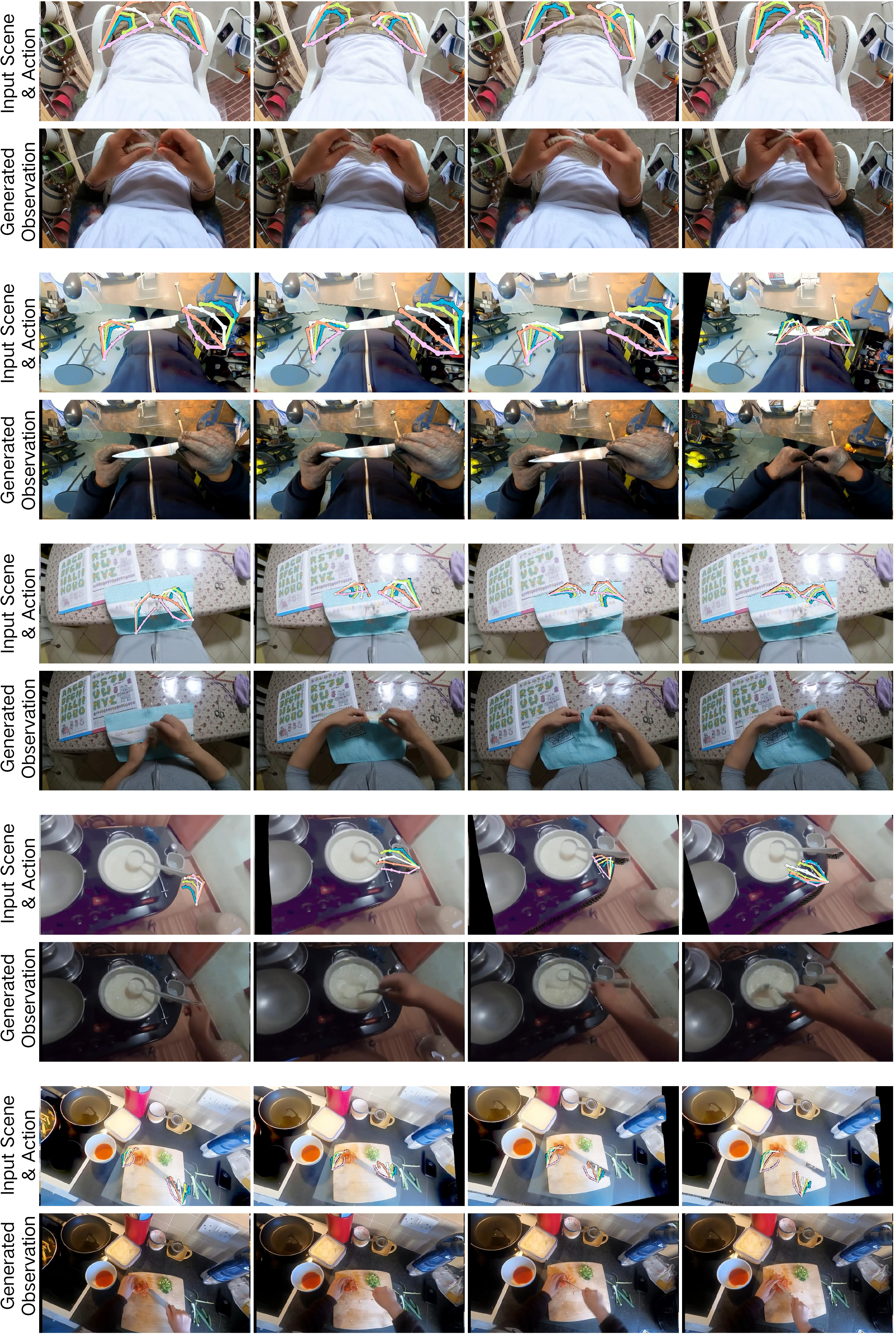}
    \caption{\textbf{Visualization} of EgoSim simulating test samples of Egovid, highlighting the ability for generating dynamic views and motions.}
    \label{fig:supp_visualize_egovid}
\end{figure*}

\begin{figure*}[t]
    \centering
    \includegraphics[width=\textwidth]{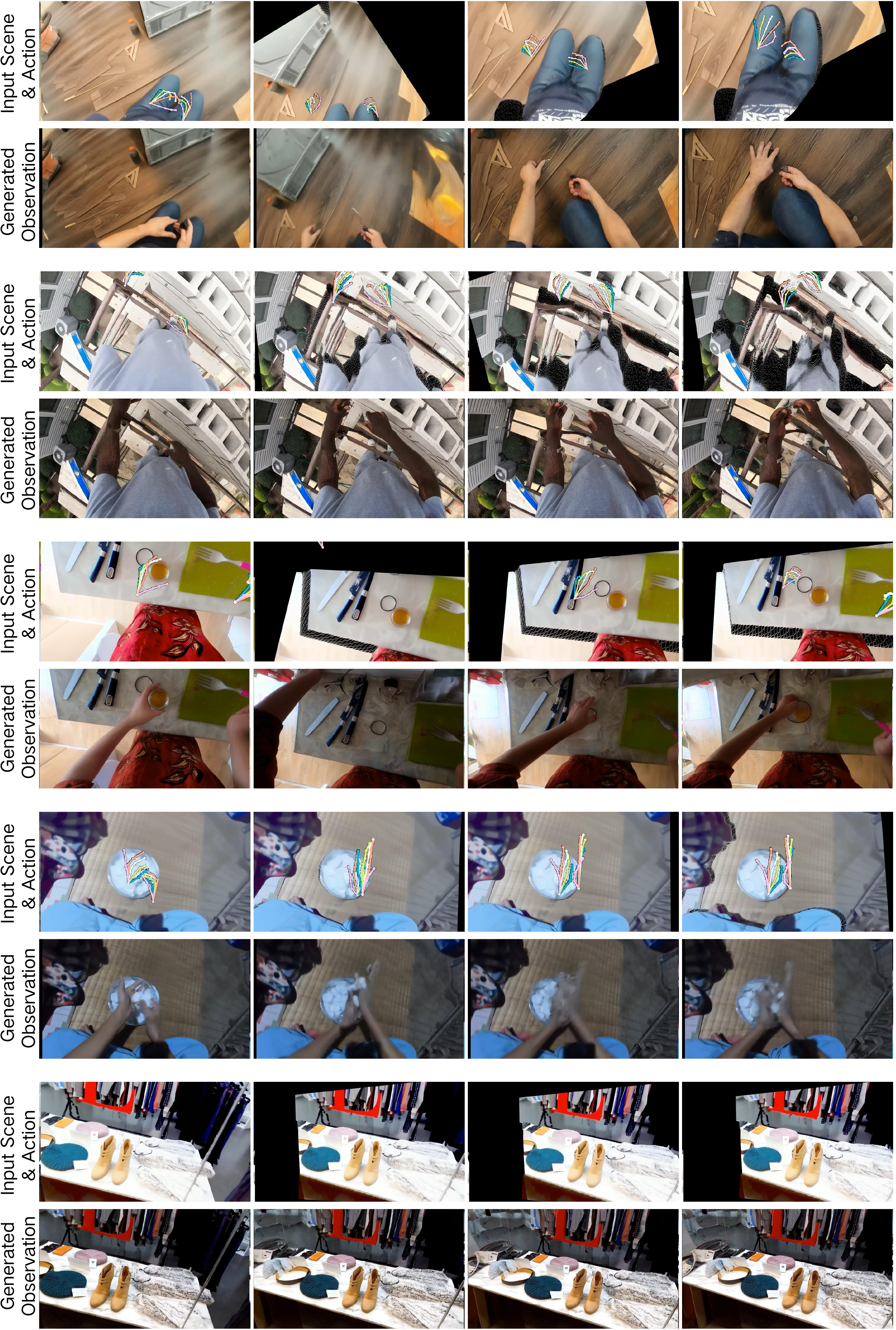}
    \caption{\textbf{Visualization} of EgoSim simulating test samples of Egovid, highlighting the ability for generating dynamic views and motions.}
    \label{fig:supp_visualize_egovid2}
\end{figure*}

\textbf{Dexterous Manipulation and Large Viewpoint Movements.} Figures~\ref{fig:supp_visualize_egodex}, \ref{fig:supp_visualize_egodex2}, \ref{fig:supp_visualize_egovid}, and \ref{fig:supp_visualize_egovid2} present our inference results on the EgoDex and EgoVid datasets. Our method exhibits precise control and high visual fidelity in fine-grained, dexterous hand-object interactions, successfully maintaining the geometric consistency of manipulated objects without introducing severe structural distortions or blending artifacts. Furthermore, despite drastic camera ego-motion and large viewpoint shifts typical of in-the-wild recordings, EgoSim maintains stable and persistent background generation. 

\begin{figure*}[t]
    \centering
    \includegraphics[width=\textwidth]{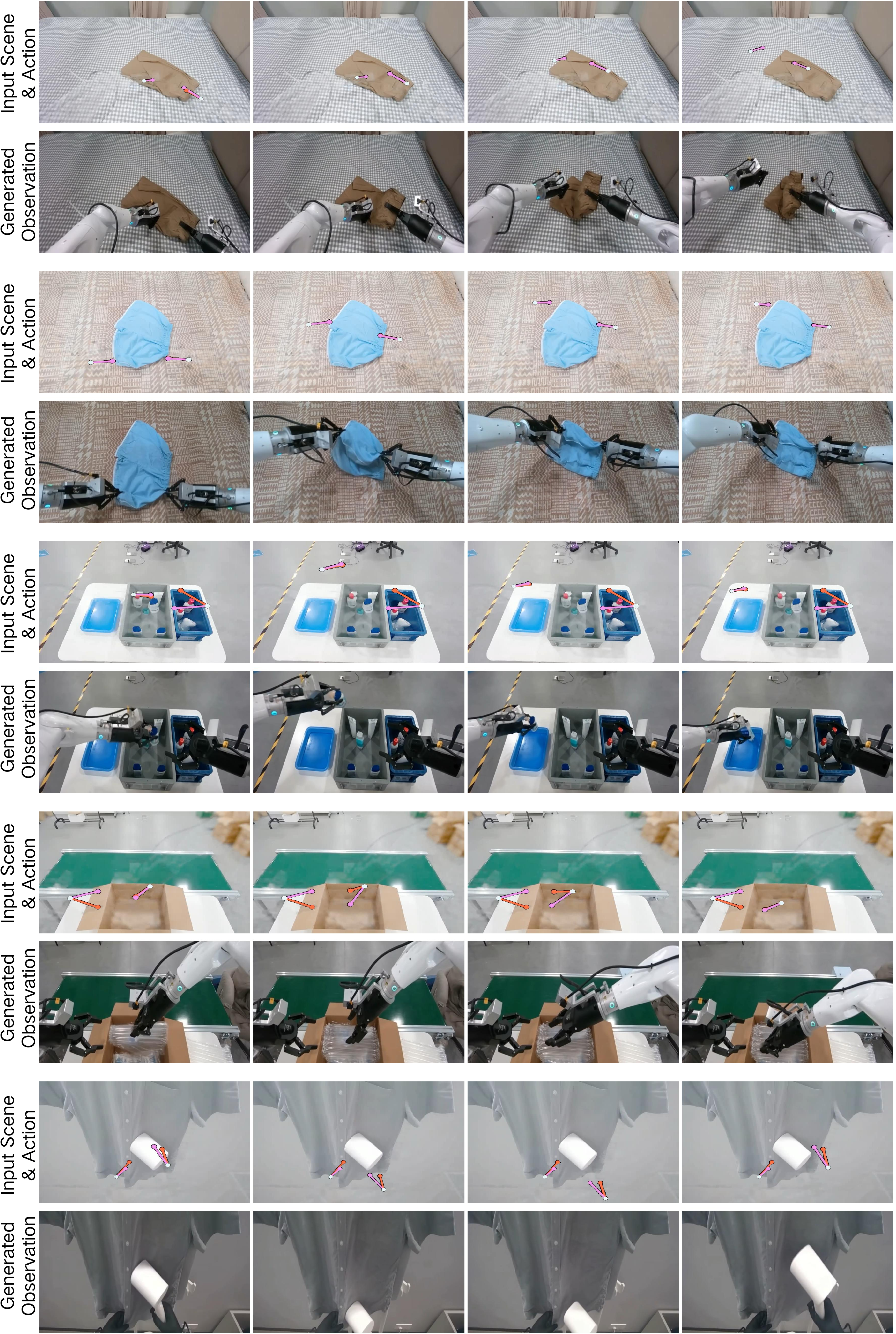}
    \caption{\textbf{Visualization} of EgoSim simulating robot interactions. }
    \label{fig:supp_visualize_agibot}
\end{figure*}

\textbf{Cross-Embodiment Manipulation.} Figure~\ref{fig:supp_visualize_agibot} presents the simulation results on the Agibot~\cite{bu2025agibot} dataset, verifying that our framework smoothly generalizes from human hands to robotic end-effectors. Guided by the extracted 3D Action keypoints, EgoSim generates physically plausible interactions between bimanual robotic grippers and target objects. Notably, our method effectively simulates the complex physical dynamics of deformable objects during robotic manipulation, correctly reflecting the state changes of items like garments and clothes under continuous interaction.

\begin{figure*}[t]
    \centering
    \includegraphics[width=\textwidth]{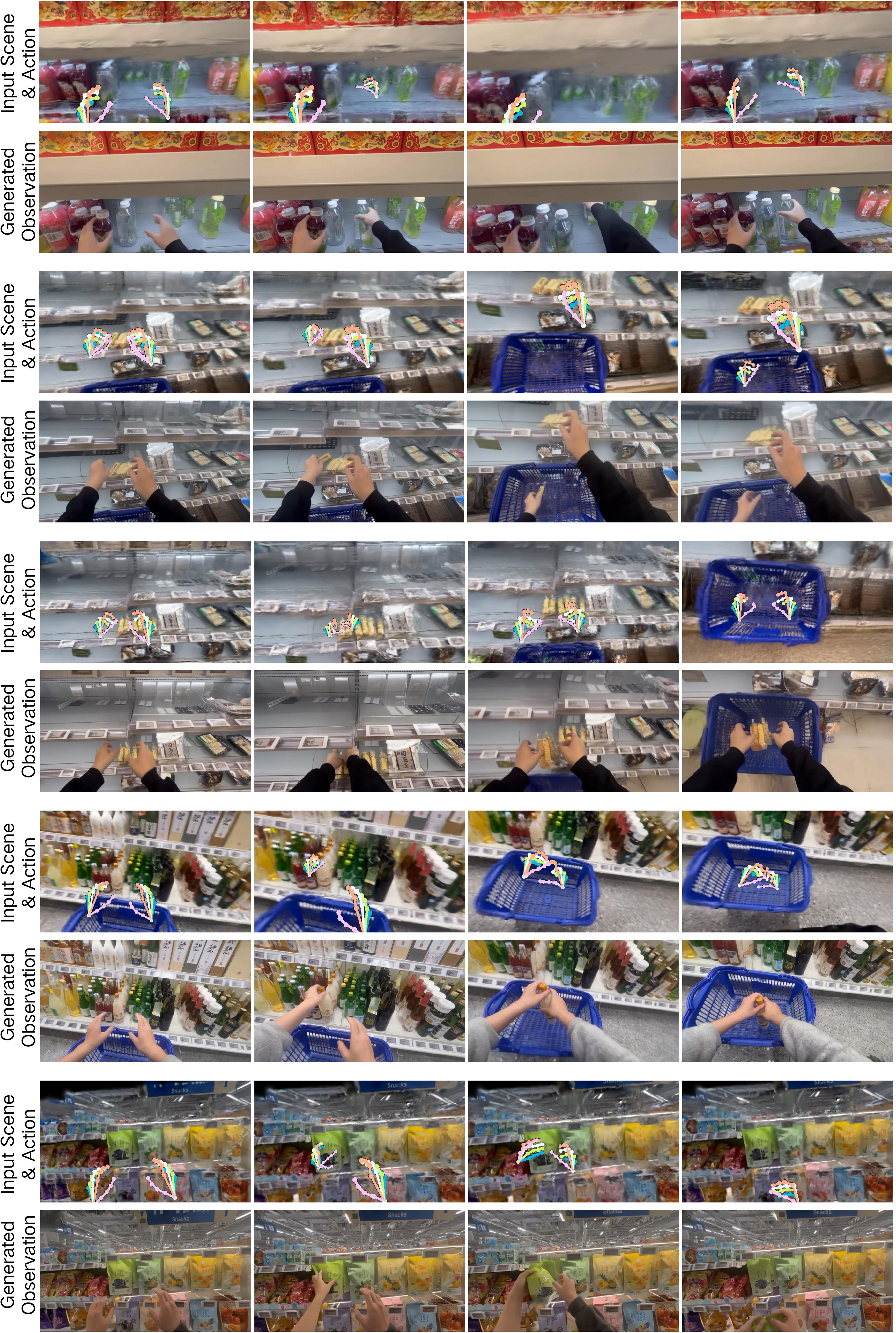}
    \caption{\textbf{Visualization} of EgoSim simulating human interactions in in-the-wild real-life scenarios.}
    \label{fig:supp_visualize_realcap}
\end{figure*}

\textbf{Real-World Object Dynamics with EgoCap.} Figure~\ref{fig:supp_visualize_realcap} illustrates the inference results on real-world scenes collected using our custom EgoCap pipeline. Tested in unstructured environments such as supermarkets, EgoSim accurately simulates realistic object dynamics, like shifting items on a shelf, while ensuring the rendering remains perfectly aligned with the complex real-world backgrounds. This confirms the robustness of both our data collection pipeline and the generative simulation model in practical, open-ended scenarios.

\clearpage

\clearpage

\clearpage

\clearpage

\clearpage

\clearpage

\clearpage

\begin{table*}[t]
\centering
\tiny
\caption{\textbf{Statistics of Task Distribution} for selected tasks from Agibot-world-beta~\cite{bu2025agibot}.}
\label{tab:task_distribution_summary}{%
\begin{tabular}{llc}
\toprule
\textbf{Task ID} & \textbf{Task Description} & \textbf{\# Clips} \\ 
\midrule
327 & Pickup items in the supermarket & 619 \\
352 & Open the fridge to get food & 3,855 \\
354 & Pickup items in the supermarket & 724 \\
356 & Packing in the supermarket & 1,105 \\
358 & Toast bread & 1,382 \\
359 & Sort in the warehouse & 1,531 \\
360 & Packing in the supermarket & 1,499 \\
361 & Flatten shorts & 15 \\
362 & Fold shorts & 25,260 \\
365 & Sort personal care products & 1,370 \\
366 & Sort food & 2,263 \\
367 & Take toast from toaster & 616 \\
374 & Sort laundry and personal care products & 124 \\
375 & Brew tea & 2,150 \\
377 & Packing in e-commerce & 4,832 \\
380 & Packing in e-commerce & 3,967 \\
384 & Insert a book into the bookshelf & 1,092 \\
385 & Pickup in the supermarket produce section & 641 \\
388 & Pickup in the supermarket & 371 \\
389 & Pickup in the supermarket & 881 \\
390 & Checkout and scan barcode in the supermarket & 4,016 \\
398 & Sort clothes & 566 \\
410 & Water pouring in restaurant & 2,302 \\
414 & Hang clothes with hanger & 11,243 \\
421 & Pick up the item to wipe away the stain & 5,631 \\
422 & Pack items for industrial logistics & 9,948 \\
424 & Clear the countertop waste & 1,363 \\
440 & Iron clothes & 6,247 \\
446 & Transport table with another robot & 293 \\
475 & Ironing clothes & 4,094 \\ 
\midrule
\textbf{Total} & & \textbf{100,000} \\ 
\bottomrule
\end{tabular}%
}
\end{table*}

\clearpage

\end{document}